\PassOptionsToPackage{table,x11names,dvipsnames,svgnames}{xcolor}
\documentclass[10pt,twocolumn,letterpaper]{article}

\usepackage[pagenumbers]{cvpr} 

\definecolor{cvprblue}{rgb}{0.21,0.49,0.74}
\usepackage[pagebackref,breaklinks,colorlinks,allcolors=cvprblue]{hyperref}

\title{A Unified Framework for Continual Learning and Unlearning}

\author{Romit Chatterjee$^{1}$\ \ \
        Vikram Chundawat$^{2}$\ \ \
        Ayush Tarun$^{3}$\ \ \
        Ankur A Mali$^{4}$\ \ \ 
        Murari Mandal$^{1}$*\\
        $^{1}$RespAI Lab, KIIT Bhubaneswar \quad $^{2}$SagepilotAI \quad  $^{3}$EPFL \quad  $^{4}$University of South Florida\\
    {\tt\small romit.respai@gmail.com} \quad 
    {\tt\small vikram@sagepilot.ai} \quad {\tt \small 
    ayush.tarun@epfl.ch} \\
   {\tt \small ankurarjunmali@usf.edu} \quad  {\tt \small murari.mandalfcs@kiit.ac.in}
}

\usepackage[hyphens]{url}  
\usepackage{graphicx} 
\usepackage{amssymb}
\usepackage{natbib}  
\usepackage{caption} 
\usepackage{algorithm}
\usepackage{algorithmic}

\usepackage{newfloat}
\usepackage{listings}
\usepackage{xspace} 
\usepackage{amsfonts}
\usepackage{amsmath}

\usepackage{tikz}

\usepackage{booktabs}
\usepackage{multirow}
\usepackage{graphicx}
\usepackage{lscape}
\usepackage{float} 
\usepackage{amsthm}
\newtheorem{theorem}{Theorem}

\newtheorem{definition}[theorem]{Definition}

\begin{document}
\maketitle
\begin{abstract}
Continual learning and machine unlearning are crucial challenges in machine learning, typically addressed separately. Continual learning focuses on adapting to new knowledge while preserving past information, whereas unlearning involves selectively forgetting specific subsets of data. In this paper, we introduce a new framework that jointly tackles both tasks by leveraging controlled knowledge distillation. Our approach enables efficient learning with minimal forgetting and effective targeted unlearning. By incorporating a fixed memory buffer, the system supports learning new concepts while retaining prior knowledge. The distillation process is carefully managed to ensure a balance between acquiring new information and forgetting specific data as needed. Experimental results on benchmark datasets show that our method matches or exceeds the performance of existing approaches in both continual learning and machine unlearning. This unified framework is the first to address both challenges simultaneously, paving the way for adaptable models capable of dynamic learning and forgetting while maintaining strong overall performance. Source code: \textcolor{blue}{https://respailab.github.io/CLMUL}
\end{abstract}
\def\thefootnote{*}
\footnotetext{Corresponding author}
\section{Introduction}
Machine learning models often face significant challenges when adapting to environments that require learning new tasks while selectively forgetting outdated or irrelevant knowledge. Continual learning~\cite{wang2024comprehensive} and machine unlearning~\cite{nguyen2022survey} are two fundamental problems that address these challenges but have largely been studied in isolation. Continual learning, also known as lifelong learning or incremental learning, focuses on enabling models to learn new information while preserving previously acquired knowledge. This is particularly challenging when new knowledge is non-stationary, and the model must adapt to evolving data distributions while mitigating catastrophic forgetting. On the other hand, machine unlearning, also referred to as forgetting or deletion, involves the selective removal of specific knowledge or data points from a model's parameters. This is crucial in scenarios involving data privacy, security, or regulatory compliance, where certain information needs to be permanently erased.\par

Despite the critical importance of both tasks, current approaches predominantly focus on either continual learning or unlearning without addressing their integration. This separation limits the development of machine learning models that can effectively handle real-world environments where both the retention of new knowledge and the selective forgetting of old information are essential.\par
\textbf{Background.} Continual learning requires a model to strike a balance between acquiring new knowledge (\textit{plasticity}) and retaining previously learned information (\textit{stability}) \cite{ororbia2022lifelong}. Various strategies have been developed to address these needs, including regularization-based methods~\cite{roady2020stream,ritter2018online,douillard2021plop}, replay-based methods~\cite{prabhu2020gdumb,li2022learning,chaudhry2019tiny,borsos2020coresets}, optimization-based approaches~\cite{chaudhry2018efficient,mirzadeh2020linear}, representation-based strategies~\cite{madaan2021representational,pham2021dualnet}, and architecture-based solutions~\cite{mallya2018piggyback}. These techniques range from imposing regularization terms to leveraging memory buffers, gradient modifications, and architectural adaptations to support continual learning in dynamic environments.

In contrast, machine unlearning focuses on forgetting specific information. This capability is vital for applications such as data privacy (e.g., removing sensitive data from models), updating models with corrected or outdated information, and minimizing the carbon footprint of GPU usage by avoiding retraining from scratch. Unlearning in deep networks has been explored primarily through probabilistic frameworks due to the non-convex nature of deep models. Recent techniques include methods based on Fisher Information Matrix~\cite{golatkar2020eternal,golatkar2020forgetting}, NTK theory~\cite{golatkar2021mixed}, and gradient update storage~\cite{yan2022arcane}. Other approaches involve error-maximizing noise~\cite{chundawat2023zero,tarun2023fast}, teacher-student frameworks~\cite{chundawat2023can,tarun2023deep,kurmanji2024towards}, and parameter attenuation at inference time~\cite{foster2024fast}. Unlearning has also been extended to generative models like text-to-image diffusion~\cite{kumari2023ablating,gandikota2023erasing} and large language models~\cite{patil2023can} using techniques such as model editing and layer unlearning.\par

\begin{figure}[t]
\centering
    \includegraphics[width=0.47\textwidth]{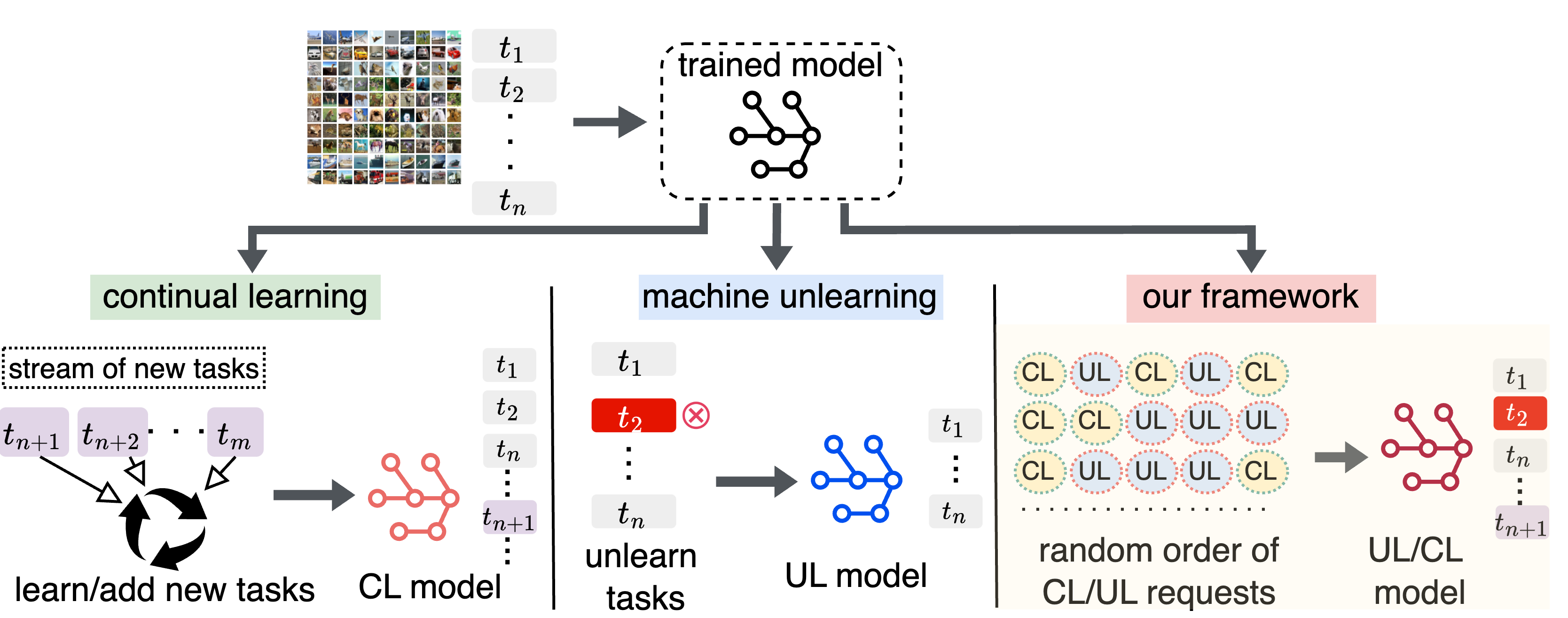}
\caption{Comparing the isolated problems of continual learning and unlearning with the unified problem of learning and unlearning in the same framework as proposed in this paper.}
\label{fig:unified_vs_others}
\end{figure}

\textbf{Motivation.} The absence of a unified framework capable of both continual learning and unlearning hinders the creation of robust and adaptable models that can dynamically learn and forget. Current disjoint approaches fail to address scenarios where a model must adapt to new knowledge while simultaneously erasing outdated or sensitive information, limiting their applicability in real-world environments. We compare the isolated continual learning, machine unlearning problems to the unified continual learning-unlearning (CL-UL) problem in Figure~\ref{fig:unified_vs_others}. In this paper, we bridge this gap by proposing a controlled knowledge distillation framework that jointly addresses both continual learning and machine unlearning. Our framework leverages knowledge distillation for selective knowledge erasure while allowing the model to integrate new concepts without losing prior information. By carefully managing the distillation process, our approach achieves a balance between the seemingly contradictory tasks of continually learning and unlearning.\par

\textbf{Our Work.} We present a knowledge distillation framework designed to handle both continual learning (CL) and unlearning (UL) operations. The framework consists of a CL teacher, a UL teacher, and a student model. For each task, the models are optimized through a unified loss function. For continual learning, two objectives ensure that previously acquired knowledge is preserved while new information is integrated. Contrastive distillation maintains existing knowledge similarities between the CL teacher and student, while adaptive distillation incorporates new learning. For unlearning, KL-Divergence between the UL teacher and student drives the targeted removal of obsolete information. The framework is carefully designed to maintain consistency even after random sequences of CL and UL requests. The CL task expands the decision space, while the UL task adjusts the space to remove specific class labels, addressing challenges related to incorrect predictions due to label removal. The replay buffer is also updated during UL to ensure privacy by deleting data related to unlearned classes.

The main contributions of this paper are summarized as follows:
\begin{itemize}
    \item \textbf{Unified Framework:} We introduce the first framework that seamlessly integrates both continual learning and machine unlearning, enabling more adaptable and robust models that can efficiently learn and forget in dynamic environments.
    
    \item \textbf{Controlled Knowledge Distillation:} We propose a distillation mechanism that selectively retains or erases knowledge depending on the task requirements, ensuring a balance between acquiring new information and forgetting outdated data.

    \item \textbf{Balancing Learning and Forgetting:} Our framework effectively manages the conflicting objectives of continual learning and unlearning, ensuring consistent performance and adaptability across a wide range of tasks.
\end{itemize}

This work not only provides a holistic solution to two critical challenges in machine learning but also lays the groundwork for future models capable of adaptive learning and forgetting, essential for real-world applications.

\section{Unified Continual Learning and Unlearning (UniCLUN)}
\textbf{Continual Learning (CL) Definition}: Given a dataset \(\mathcal{D} = \{(\mathbf{x}_i, y_i)\}_{i=1}^N\) with inputs \(\mathbf{x}_i \in \mathbb{R}^d\) and labels \(y_i \in \mathbb{R}\), and a model \(\mathcal{M}\) with parameters \(\theta \in \mathbb{R}^p\), continual learning involves updating \(\mathcal{M}\) sequentially on datasets \(\{\mathcal{D}_{1}, \mathcal{D}_{2}, \ldots, \mathcal{D}_{K}\}\). For each new dataset \(\mathcal{D}_k\), the model learns to map test inputs \(\mathbf{x}\) from any previous task \(t_k\) to the correct class \(y\).\\
\textbf{Machine Unlearning (UL) Definition}: For unlearning, let \(\mathcal{D} = \mathcal{D}_r \cup \mathcal{D}_f\), where \(\mathcal{D}_f\) is the subset of data to forget, and \(\mathcal{D}_r\) is the data to retain. If \(\mathcal{M}'\) denotes the model retrained on \(\mathcal{D}_r\), then the unlearning objective is to adjust \(\theta\) to produce parameters \(\theta_u\) in a model \(\mathcal{M}_u\) such that:
\begin{equation}
\label{eq:unlearn_similarity}
    \mathcal{P}(\mathcal{M}_u(x, \theta_u) = y) \approx \mathcal{P}(\mathcal{M}'(x, \theta') = y), \quad \forall x \in \mathcal{D}, \; y \in \mathbb{R}
\end{equation}
where \(\mathcal{P}(X)\) denotes the probability distribution of any random variable \(X\).\\
\textbf{Unified Continual Learning and Unlearning (UniCLUN)}: The UniCLUN framework addresses both continual learning and unlearning requests that may arrive unpredictably, requiring flexible adaptation to either task.\par

\begin{definition}
In UniCLUN, each task \(t\) may be a continual learning (CL) task (learning new data) or an unlearning (UL) task (removing specified knowledge). For CL tasks, \(\mathcal{M}\) is updated to map test inputs \(\mathbf{x}\) from any learned task \(t_k\) (\(k = 1, \ldots, K\)) to its correct label \(y\). For UL tasks, \(\mathcal{M}_u\) satisfies Eq.~\ref{eq:unlearn_similarity}, ensuring that knowledge removal does not degrade retained task performance.
\end{definition}
UniCLUN balances learning and selective forgetting within a unified structure, essential for dynamic environments where models must adapt continually while preserving task integrity and accuracy.

\begin{figure}[]
\centering
    \includegraphics[width=0.5\textwidth]{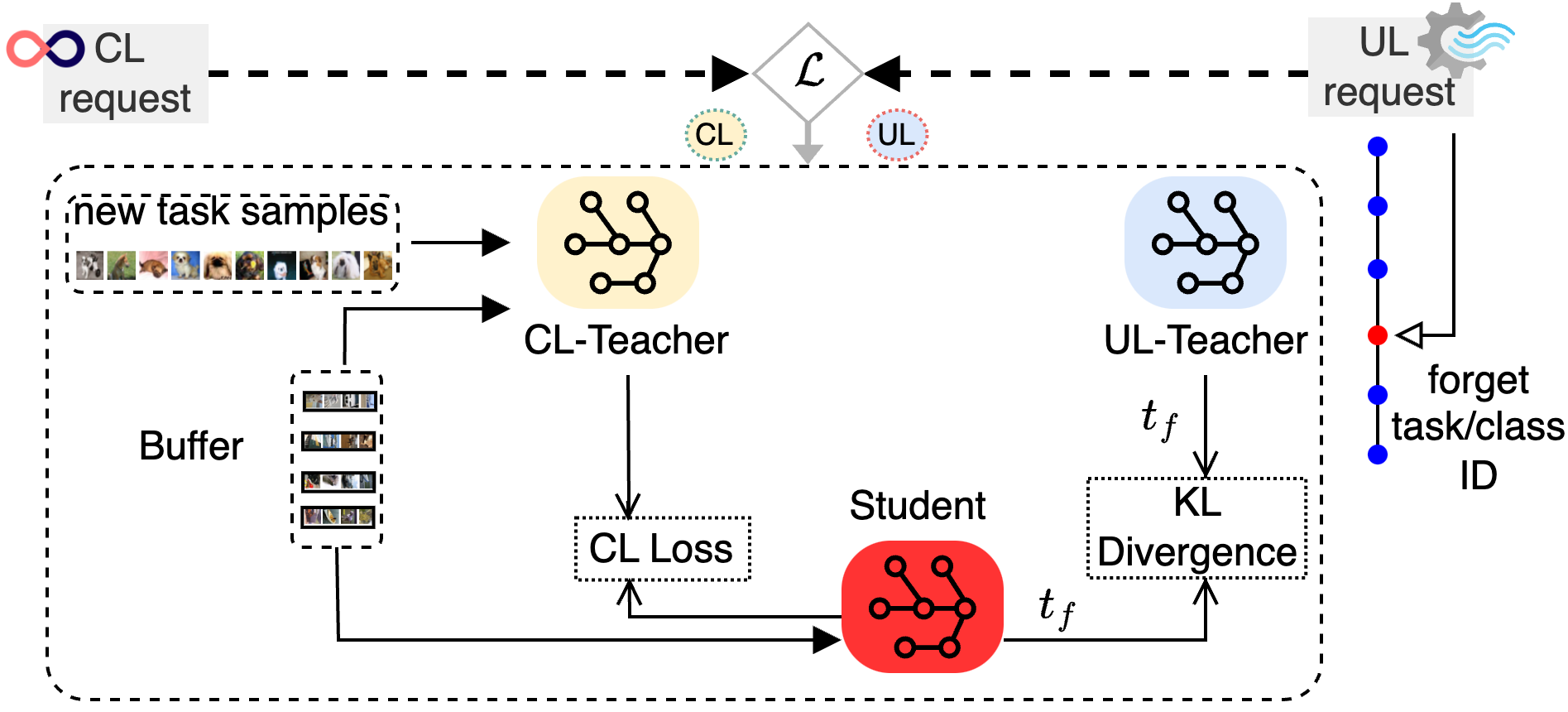}
\caption{The proposed controlled knowledge distillation based unified framework for continual learning and unlearning.}
\label{fig:proposed_framework}
\end{figure}

\section{Proposed UniCLUN Method}
We introduce \textbf{UniCLUN}, a unified framework for \textit{continual learning} (CL) and \textit{unlearning} (UL) that employs a \textit{teacher-student contrastive distillation} strategy. UniCLUN is designed to meet the dual requirements of continuously integrating new knowledge and selectively forgetting information upon request, ensuring relevance and adaptability of learned knowledge. An overview of our proposed method is illustrated in Figure~\ref{fig:proposed_framework}. Our framework leverages a multi-teacher, single-student network architecture for effective knowledge distillation in both continual learning and unlearning tasks.

While knowledge distillation has been explored individually for continual learning~\cite{cha2021co2l,buzzega2020dark,li2022learning} and machine unlearning~\cite{chundawat2023can,tarun2023deep,kurmanji2024towards}, the unified solution presented here introduces specific adaptations for managing these tasks concurrently. By employing a shared student model that receives guidance from multiple teacher models, UniCLUN efficiently consolidates new information and removes obsolete or unwanted data, offering a cohesive and resource-effective alternative to handling these tasks separately.
The architecture consists of two primary modules: the \textbf{Continual Learning (CL) Module} and the \textbf{Unlearning (UL) Module}.

\subsection{Continual Learning (CL) Module} 

To support continual learning, we incorporate a \textbf{bounded replay buffer} \(\mathcal{B} = \{(x_i, y_i)_{i=1}^{|\mathcal{B}|}\}\) of size \(|\mathcal{B}|\), which stores a subset of samples from previously encountered tasks. This buffer is crucial in retaining knowledge by providing the model with reference samples during CL tasks, aiding in preventing knowledge erosion. For UL tasks, the buffer ensures that samples associated with forgotten data are also removed, maintaining the integrity of unlearned information. The replay buffer is dynamically updated to support smooth transitions between continual learning and unlearning. We employ a reservoir sampling strategy~\cite{vitter1985random} to selectively add and delete samples, ensuring that the buffer remains compact and relevant, maximizing the effectiveness of knowledge retention and unlearning as required. By effectively coordinating these modules within a unified framework, UniCLUN presents a versatile and efficient approach to simultaneously handling continual learning and unlearning tasks. This unified strategy avoids the redundancy and inefficiency of separate systems, providing a streamlined, adaptable solution suitable for complex and evolving learning environments.

We propose a robust framework for continual learning that integrates both a \textbf{student model} and a \textbf{teacher model}, initially set with identical parameters. The overarching goal is for the teacher model, through selective knowledge retention and adaptation to new tasks, to serve as a reliable model for deployment. This framework allows the teacher model to effectively manage both learning new information and unlearning specified knowledge. A momentum update mechanism is employed, where the student model’s parameters are intermittently incorporated into the teacher model, enabling the accumulation of new knowledge while retaining core information from previously learned tasks. Each model is composed of three main components: a feature extractor \( f_{\Theta} \) for encoding feature representations, a classifier \( f_{\Phi} \) for mapping feature representations to output labels, and a projector \( f_{\Psi} \) for embedding features into a latent space where contrastive distillation is applied. This structured architecture enables precise, targeted knowledge transfer between the student and teacher models.\par

\textbf{Classification loss for knowledge acquisition.} The initial optimization involves training the student model on both new task samples and a replay buffer containing samples from previous tasks. This process is guided by a classification loss \(\mathcal{L}_{ce}\), which is designed to encourage accurate predictions on both new and previously encountered data. Formally, the classification loss is defined as:

\begin{equation}
    \mathcal{L}_{ce} = \mathbb{E}_{(x, y) \sim \mathcal{D}_t \cup \mathcal{B}} \ell(f_{\Theta_s, \Phi_s}(x), y),
\end{equation}
where \( f_{\Theta_s, \Phi_s} \) represents the combined output of the feature extractor \( f_{\Theta_s} \) and classifier \( f_{\Phi_s} \) in the student model, \(\mathcal{D}_t\) is the current task dataset, and \(\mathcal{B}\) is the replay buffer.\par

\textbf{Online distillation loss for knowledge retention.} To mitigate catastrophic forgetting, a \textbf{weighted distillation loss} is introduced to retain essential knowledge from prior tasks while learning new classifications. The weighting factor \(\omega(x_i)\) dynamically scales the importance of each sample based on the teacher model's confidence in the sample’s class label. This weight is computed as:
\begin{equation}
    \omega(x_i) = \frac{\exp(f_{\Theta_T, \Phi_T}(x_i)_{y_i} / \rho)}{\sum_{c' = 1}^{C} \exp(f_{\Theta_T, \Phi_T}(x_i)_{c'} / \rho)},
\end{equation}
where \( f_{\Theta_T, \Phi_T} \) represents the teacher’s combined feature extractor and classifier output, \(\rho\) is a temperature parameter that controls sharpness, and \(C\) is the total number of classes. The online distillation loss \(\mathcal{L}_{od}\) is defined as:
\begin{equation}
    \mathcal{L}_{od} = \mathbb{E}_{x_i \sim \mathcal{B}} \left[\omega(x_i) \| f_{\Theta_T, \Phi_T}(x_i) - f_{\Theta_s, \Phi_s}(x_i) \|^2_2 \right].
\end{equation}
This loss term encourages alignment between the teacher and student predictions, thereby consolidating previous knowledge within the student model while learning new data.\par

\textbf{Contrastive distillation for embedding alignment.} In addition to distilling the teacher’s responses, we apply a \textbf{contrastive distillation loss} to ensure that the teacher and student models maintain similar embeddings in the latent space, enhancing the consistency of learned representations. Let \( z_T = f_{\Theta_T, \Psi_T}(x) \) and \( z_s = f_{\Theta_s, \Psi_s}(x) \) represent the embeddings produced by the teacher and student models, respectively, after combining the feature extractor \( f_{\Theta} \) and projector \( f_{\Psi} \). The contrastive distillation loss \(\mathcal{L}_{cd}\) is defined as:
\begin{equation}
    \mathcal{L}_{cd} = \sum_{z_j^T \sim z^{T+}} \log \frac{h(z_i^s, z_j^T)}{\sum_{z_k^T \sim z_T} h(z_i^s, z_k^T)},
\end{equation}
where \( z^{T+} \) denotes the set of teacher embeddings with the same label as \( z_i^s \), and \( h \) is a critic function indicating joint distribution membership, defined as:
\begin{equation}
    h(z_i, z_j) = \frac{\exp\left( \frac{(z_i / \|z_i\|_2)^{\intercal} (z_j / \|z_j\|_2)}{\tau} \right)}{\exp(1 / \tau)},
\end{equation}
where \(\tau\) is a temperature hyperparameter, and \((\cdot)^{\intercal}\) denotes the transpose operation.\par

\textbf{Supervised contrastive distillation for class similarity.} To further enhance class-wise alignment, we introduce a \textbf{supervised contrastive distillation loss} \(\mathcal{L}_{scd}\) that encourages intra-class similarity between student embeddings. This loss is given by:
\begin{equation}
    \mathcal{L}_{scd} = -\mathbb{E}_{z_i^s \sim z^s} \sum_{z_j^s \sim z^{s+}} \log \frac{h(z_i^s, z_j^s)}{\sum_{z_k^s \sim z^s} h(z_i^s, z_k^s)},
\end{equation}
where \( z^{s+} \) represents the set of student embeddings with the same label as \( z_i^s \). By minimizing \(\mathcal{L}_{cd}\) and \(\mathcal{L}_{scd}\) together, the student model effectively consolidates previously learned knowledge while acquiring new tasks.\\

\textbf{Overall continual learning objective.} The total continual learning objective \(\mathcal{L}_{cl}\) combines these loss terms, enabling the student model to balance knowledge retention and acquisition. It is defined as:
\begin{equation}
    \mathcal{L}_{cl} = \mathcal{L}_{ce} + \alpha_1 \mathcal{L}_{od} + \alpha_2 \mathcal{L}_{cd} + \alpha_3 \mathcal{L}_{scd},
\end{equation}
where \(\alpha_1\), \(\alpha_2\), and \(\alpha_3\) are hyperparameters that control the contribution of each loss term, allowing for a flexible yet robust training process. Table \ref{tab:loss_removal_comparison} shows contributions of each loss.\\

\subsection{Unlearning (UL) Module}

To selectively remove specific task knowledge, we use a \textit{bad teacher} model \( f_{\Theta_b, \Phi_b} \) with parameters \(\Theta_b\) and \(\Phi_b\) for the feature extractor and classifier, respectively. This model is initialized without the task knowledge intended for removal, complementing the original teacher model \( f_{\Theta_T, \Phi_T} \) from the continual learning setup. The original teacher reinforces knowledge retention on desired data, while the bad teacher guides the student model in selectively unlearning task-specific information. In this configuration, both teachers distill knowledge into the student model to achieve selective unlearning. For unlearning task samples, the bad teacher serves as the primary distillation source, while buffer samples (retained knowledge) are distilled from the original teacher. Additionally, any unlearned task samples are removed from the buffer to maintain consistency. The continual unlearning objective, \(\mathcal{L}_{cu}\), is defined as:
\begin{equation}
\begin{split}
    \mathcal{L}_{cu} = & \, (1 - \omega_u) \cdot \mathcal{KL}(f_{\Theta_T, \Phi_T}(x) \| f_{\Theta_s, \Phi_s}(x)) \\
    & + \omega_u \cdot \mathcal{KL}(f_{\Theta_b, \Phi_b}(x) \| f_{\Theta_s, \Phi_s}(x)),
\end{split}
\end{equation}
where \(\omega_u\) is a dynamically adjusted weight that prioritizes the original teacher for buffer samples and the bad teacher for unlearning samples. Here, \(\mathcal{KL}(p \| q)\) denotes the KL-Divergence:
\begin{equation}
    \mathcal{KL}(p\|q) = \sum_{i} p^{(i)} \log\left(\frac{p^{(i)}}{q^{(i)}}\right).
\end{equation}
This setup enables the selective removal of task-specific information while preserving the model’s performance on retained tasks.\par

\subsection{Unified Training Objective} 
The unified objective integrates both continual learning (CL) and unlearning (UL) tasks within a single adaptive framework, eliminating the need for explicit task flags. The overall objective is defined as:
\begin{equation}
    \mathcal{L} = \gamma \cdot \mathcal{L}_{cl} + (1 - \gamma) \cdot \mathcal{L}_{cu},
\end{equation}
where \(\gamma\) is a context-sensitive weighting factor that adjusts based on task requirements, allowing seamless transitions between learning and unlearning. When prioritizing knowledge acquisition, \(\gamma\) is set closer to 1, focusing on \(\mathcal{L}_{cl}\), while for unlearning tasks, \(\gamma\) shifts towards 0, emphasizing \(\mathcal{L}_{cu}\). This unified objective allows the model to adapt continuously to task requirements without rigid switching, maintaining efficient performance across diverse learning and unlearning demands. By dynamically balancing \(\mathcal{L}_{cl}\) and \(\mathcal{L}_{cu}\) through \(\gamma\), the model achieves robust adaptation to complex task sequences while preserving computational efficiency and stability.\par

\textbf{Teacher model update with contextualized momentum.} Following the student model’s optimization, the teacher model undergoes a contextually-driven momentum update, enhancing alignment between the teacher and the student parameters. This update is achieved through the following equations:
\[
\Theta_T \leftarrow m\Theta_T + (1 - m)\left[(1 - X)\Theta_T + X\Theta_s\right]
\]
\[
\Phi_T \leftarrow m\Phi_T + (1 - m)\left[(1 - X)\Phi_T + X\Phi_s\right]
\]
\[
\Psi_T \leftarrow m\Psi_T + (1 - m)\left[(1 - X)\Psi_T + X\Psi_s\right]
\]
where, \( m \) is the momentum coefficient and \( X \) is a random variable with a Bernoulli distribution:
\[
P(X = k) = p^k(1 - p)^{1 - k}, \quad k \in \{0, 1\}
\]
This adaptive update approach introduces stochastic variability, allowing the teacher model to incorporate relevant changes based on the student’s contextual learning dynamics, thus refining the model’s performance for varied tasks. In the supplementary material we conduct detailed ablation study to show significance of each hyperparameters. 

\begin{table}[t]
    \centering
    \resizebox{\linewidth}{!}{
    \begin{tabular}{c c c c c c c c c c c c c}
        \toprule
        \multirow{2}{*}{BS} & Execution & \multicolumn{2}{c}{Task 1} & \multicolumn{2}{c}{Task 2} & \multicolumn{2}{c}{Task 3} & \multicolumn{2}{c}{Task 4} & \multicolumn{2}{c}{Task 5} \\ 
        \cmidrule{3-12}
         & Sequence & C0 & C1 & C2 & C3 & C4 & C5 & C6 & C7 & C8 & C9 \\ \toprule
        \multirow{6}{*}{\rotatebox[origin=c]{-90}{\centering 200}} & Learn T1 & \cellcolor[HTML]{EFEFEF}\textbf{99.4} & \cellcolor[HTML]{EFEFEF}\textbf{99.3} & 0.0 & 0.0 & 0.0 & 0.0 & 0.0 & 0.0 & 0.0 & 0.0 \\ 
         & Learn T2 & 53.6 & 94.0 & \cellcolor[HTML]{EFEFEF}\textbf{96.0} & \cellcolor[HTML]{EFEFEF}\textbf{95.0} & 0.0 & 0.0 & 0.0 & 0.0 & 0.0 & 0.0 \\ 
         & Unlearn T2 & 97.8 & 98.5 & \cellcolor[HTML]{EFEFEF}\textbf{0.1} & \cellcolor[HTML]{EFEFEF}\textbf{0.2} & 0.0 & 0.0 & 0.0 & 0.0 & 0.0 & 0.0 \\ 
         & Learn T3 & 57.4 & 88.4 & 0.2 & 0.1 & \cellcolor[HTML]{EFEFEF}\textbf{98.2} & \cellcolor[HTML]{EFEFEF}\textbf{94.9} & 0.0 & 0.0 & 0.0 & 0.0 \\ 
         & Learn T4 & 59.6 & 88.1 & 0.1 & 0.1 & 39.0 & 43.2 & \cellcolor[HTML]{EFEFEF}\textbf{98.2} & \cellcolor[HTML]{EFEFEF}\textbf{99.0} & 0.0 & 0.0 \\ 
         & Learn T5 & 24.4 & 45.6 & 0.3 & 0.1 & 55.0 & 60.8 & 85.4 & 69.6 & \cellcolor[HTML]{EFEFEF}\textbf{98.7} & \cellcolor[HTML]{EFEFEF}\textbf{99.0} \\ \midrule
        \multirow{6}{*}{\rotatebox[origin=c]{-90}{\centering 500}} & Learn T1 & \cellcolor[HTML]{EFEFEF}\textbf{99.4} & \cellcolor[HTML]{EFEFEF}\textbf{99.1} & 0.0 & 0.0 & 0.0 & 0.0 & 0.0 & 0.0 & 0.0 & 0.0 \\ 
         & Learn T2 & 79.6 & 93.5 & \cellcolor[HTML]{EFEFEF}\textbf{94.6} & \cellcolor[HTML]{EFEFEF}\textbf{96.4} & 0.0 & 0.0 & 0.0 & 0.0 & 0.0 & 0.0 \\ 
         & Unlearn T2 & 99.4 & 97.3 & \cellcolor[HTML]{EFEFEF}\textbf{0.2} & \cellcolor[HTML]{EFEFEF}\textbf{0.3} & 0.0 & 0.0 & 0.0 & 0.0 & 0.0 & 0.0 \\ 
         & Learn T3 & 93.0 & 95.6 & 0.1 & 0.2 & \cellcolor[HTML]{EFEFEF}\textbf{97.4} & \cellcolor[HTML]{EFEFEF}\textbf{96.3} & 0.0 & 0.0 & 0.0 & 0.0 \\ 
         & Learn T4 & 81.0 & 91.7 & 0.1 & 0.1 & 43.3 & 54.1 & \cellcolor[HTML]{EFEFEF}\textbf{99.1} & \cellcolor[HTML]{EFEFEF}\textbf{98.8} & 0.0 & 0.0 \\ 
         & Learn T5 & 58.1 & 73.0 & 0.1 & 0.1 & 60.5 & 67.8 & 75.2 & 74.0 & \cellcolor[HTML]{EFEFEF}\textbf{98.7} & \cellcolor[HTML]{EFEFEF}\textbf{99.1} \\ \midrule
        \multirow{6}{*}{\rotatebox[origin=c]{-90}{\centering 5120}} & Learn T1 & \textbf{99.4} & \cellcolor[HTML]{EFEFEF}\textbf{98.9} & 0.0 & 0.0 & 0.0 & 0.0 & 0.0 & 0.0 & 0.0 & 0.0 \\ 
         & Learn T2 & 95.2 & 99.1 & \cellcolor[HTML]{EFEFEF}\textbf{91.2} & \cellcolor[HTML]{EFEFEF}\textbf{93.6} & 0.0 & 0.0 & 0.0 & 0.0 & 0.0 & 0.0 \\ 
         & Unlearn T2 & 99.8 & 98.9 & \cellcolor[HTML]{EFEFEF}\textbf{0.1} & \cellcolor[HTML]{EFEFEF}\textbf{0.1} & 0.0 & 0.0 & 0.0 & 0.0 & 0.0 & 0.0 \\ 
         & Learn T3 & 98.4 & 97.9 & 0.2 & 0.1 & \cellcolor[HTML]{EFEFEF}\textbf{97.0} & \cellcolor[HTML]{EFEFEF}\textbf{96.7} & 0.0 & 0.0 & 0.0 & 0.0 \\ 
         & Learn T4 & 97.2 & 97.7 & 0.1 & 0.1 & 89.6 & 85.8 & \cellcolor[HTML]{EFEFEF}\textbf{95.9} & \cellcolor[HTML]{EFEFEF}\textbf{96.7} & 0.0 & 0.0 \\ 
         & Learn T5 & 91.4 & 94.0 & 0.1 & 0.1 & 89.2 & 88.2 & 95.2 & 91.1 & \cellcolor[HTML]{EFEFEF}\textbf{96.1} & \textbf{96.8}\\ \bottomrule
    \end{tabular}}
    \caption{CL and single task UL in CIFAR-10 in a $2\times5$ task distribution setup. UL of Task 2 can be observed with accuracy dropping to $\sim0.1\%-0.3\%$ for the corresponding classes. Similarly CL accuracy gains in new Tasks are highlighted with~\textbf{bold}.}
    \label{tab:cifar10_2x5_1}
    \vspace{-1\baselineskip}
\end{table}

\begin{table}[t]
    \centering
    \resizebox{\linewidth}{!}{
    \begin{tabular}{c c c c c c c c c c c c c}
        \toprule
        \multirow{2}{*}{BS} & Execution & \multicolumn{2}{c}{Task 1} & \multicolumn{2}{c}{Task 2} & \multicolumn{2}{c}{Task 3} & \multicolumn{2}{c}{Task 4} & \multicolumn{2}{c}{Task 5} \\ \cmidrule{3-12}
         & Sequence & C0 & C1 & C2 & C3 & C4 & C5 & C6 & C7 & C8 & C9 \\ \toprule
        \multirow{4}{*}{\rotatebox[origin=c]{-90}{\centering 200}} & Learn initial 4 tasks & \cellcolor[HTML]{EFEFEF}\textbf{70.0} & \cellcolor[HTML]{EFEFEF}\textbf{79.5} & \cellcolor[HTML]{EFEFEF}\textbf{35.0} & \cellcolor[HTML]{EFEFEF}\textbf{26.6} & \cellcolor[HTML]{EFEFEF}\textbf{53.3} & \cellcolor[HTML]{EFEFEF}\textbf{65.2} & \cellcolor[HTML]{EFEFEF}\textbf{98.7} & \cellcolor[HTML]{EFEFEF}\textbf{97.8} & 0.0 & 0.0 \\ 
         & Unlearn T2 & 73.6 & 86.5 & \cellcolor[HTML]{EFEFEF}\textbf{0.1} & \cellcolor[HTML]{EFEFEF}\textbf{0.1} & 93.7 & 88.8 & 98.5 & 91.1 & 0.0 & 0.0 \\ 
         & Unlearn T4 & 75.6 & 83.4 & 0.1 & 0.2 & 97.6 & 94.5 & \cellcolor[HTML]{EFEFEF}\textbf{0.3} & \cellcolor[HTML]{EFEFEF}\textbf{0.4} & 0.0 & 0.0 \\ 
         & Learn T5 & 19.6 & 39.4 & 0.1 & 0.1 & 81.8 & 83.4 & 0.3 & 0.1 & \cellcolor[HTML]{EFEFEF}\textbf{98.1} & \cellcolor[HTML]{EFEFEF}\textbf{99.8} \\ \midrule
        \multirow{4}{*}{\rotatebox[origin=c]{-90}{\centering 500}} & Learn initial 4 tasks & \cellcolor[HTML]{EFEFEF}\textbf{75.2} & \cellcolor[HTML]{EFEFEF}\textbf{98.5} & \cellcolor[HTML]{EFEFEF}\textbf{53.7} & \cellcolor[HTML]{EFEFEF}\textbf{40.8} & \cellcolor[HTML]{EFEFEF}\textbf{81.9} & \cellcolor[HTML]{EFEFEF}\textbf{68.8} & \cellcolor[HTML]{EFEFEF}\textbf{98.5} & \cellcolor[HTML]{EFEFEF}\textbf{97.8} & 0.0 & 0.0 \\ 
         & Unlearn T2 & 90.4 & 97.5 & \cellcolor[HTML]{EFEFEF}\textbf{0.3} & \cellcolor[HTML]{EFEFEF}\textbf{0.1} & 94.7 & 93.7 & 97.8 & 91.6 & 0.0 & 0.0 \\ 
         & Unlearn T4 & 88.2 & 96.0 & 0.1 & 0.2 & 98.6 & 93.7 & \cellcolor[HTML]{EFEFEF}\textbf{4.0} & \cellcolor[HTML]{EFEFEF}\textbf{1.8} & 0.0 & 0.0 \\ 
         & Learn T5 & 47.8 & 65.9 & 0.2 & 0.3 & 90.2 & 88.0 & 1.0 & 0.3 & \cellcolor[HTML]{EFEFEF}\textbf{99.1} & \cellcolor[HTML]{EFEFEF}\textbf{99.2} \\ \midrule
        \multirow{4}{*}{\rotatebox[origin=c]{-90}{\centering 5120}} & Learn initial 4 tasks & \cellcolor[HTML]{EFEFEF}\textbf{97.2} & \cellcolor[HTML]{EFEFEF}\textbf{99.4} & \cellcolor[HTML]{EFEFEF}\textbf{84.0} & \cellcolor[HTML]{EFEFEF}\textbf{74.3} & \cellcolor[HTML]{EFEFEF}\textbf{90.2} & \cellcolor[HTML]{EFEFEF}\textbf{81.6} & \cellcolor[HTML]{EFEFEF}\textbf{95.2} & \cellcolor[HTML]{EFEFEF}\textbf{94.5} & 0.0 & 0.0 \\ 
         & Unlearn T2 & 99.2 & 98.7 & \cellcolor[HTML]{EFEFEF}\textbf{23.6} & \cellcolor[HTML]{EFEFEF}\textbf{8.7} & 96.6 & 91.9 & 95.5 & 96.1 & 0.0 & 0.0 \\ 
         & Unlearn T4 & 99.4 & 98.9 & 30.4 & 2.3 & 97.8 & 96.7 & \cellcolor[HTML]{EFEFEF}\textbf{2.0} & \cellcolor[HTML]{EFEFEF}\textbf{0.9} & 0.0 & 0.0 \\ 
         & Learn T5 & 84.8 & 90.2 & 0.2 & 0.2 & 96.0 & 95.1 & 0.3 & 0.1 & \cellcolor[HTML]{EFEFEF}\textbf{98.7} & \cellcolor[HTML]{EFEFEF}\textbf{98.8} \\ \bottomrule
    \end{tabular}}
    \caption{CL and multiple task UL in CIFAR-10 in a $2\times5$ task distribution. UL of Task 2 and Task 4 can be observed with accuracy dropping to $\sim 0.1\%-4\%$ for the corresponding classes. Similarly CL accuracy gains in new Tasks are highlighted with~\textbf{bold}.}
    \label{tab:cifar10_2x5_4}
    \vspace{-1\baselineskip}
\end{table}

\begin{table}[t]
\centering
\resizebox{\linewidth}{!}{
\begin{tabular}{l l l l l l l l l l l l}
\toprule
\multirow{2}{*}{BS} & Execution & T1 & T2 & T3 & T4 & T5 & T6 & T7 & T8 & T9 & T10 \\ \cmidrule{3-12}
 &  Sequence & C-0 & C-1 & C-2 & C-3 & C-4 & C-5 & C-6 & C-7 & C-8 & C-9 \\
\toprule
\multirow{9}{*}{\rotatebox[origin=c]{-90}{\centering 200}} & Learn first 5 tasks & \cellcolor[HTML]{EFEFEF}\textbf{37.2} & \cellcolor[HTML]{EFEFEF}\textbf{26.8} & \cellcolor[HTML]{EFEFEF}\textbf{17.1} & \cellcolor[HTML]{EFEFEF}\textbf{10.9} & \textbf{99.8} & 0.0 & 0.0 & 0.0 & 0.0 & 0.0 \\
 & Unlearn T2 & 57.8 & \cellcolor[HTML]{EFEFEF}\textbf{0.1} & 42.4 & 69.1 & 91.6 & 0.0 & 0.0 & 0.0 & 0.0 & 0.0 \\
 & Unlearn T4 & 59.0 & 0.2 & 85.1 & \cellcolor[HTML]{EFEFEF}\textbf{6.3} & 82.2 & 0.0 & 0.0 & 0.0 & 0.0 & 0.0 \\
 & Unlearn T5 & 57.2 & 0.1 & 94.2 & 0.7 & \cellcolor[HTML]{EFEFEF}\textbf{34.5} & 0.0 & 0.0 & 0.0 & 0.0 & 0.0 \\
 & Learn T6 & 35.8 & 0.2 & 44.7 & 0.3 & 0.2 & \cellcolor[HTML]{EFEFEF}\textbf{99.3} & 0.0 & 0.0 & 0.0 & 0.0 \\
 & Learn T7 & 49.0 & 0.2 & 23.8 & 0.7 & 0.2 & 9.4 & \cellcolor[HTML]{EFEFEF}\textbf{99.7} & 0.0 & 0.0 & 0.0 \\
 & Learn T8 & 41.8 & 0.1 & 30.2 & 0.3 & 0.2 & 6.2 & 47.4 & \cellcolor[HTML]{EFEFEF}\textbf{99.6} & 0.0 & 0.0 \\
 & Learn T9 & 9.6 & 0.2 & 33.7 & 0.1 & 0.1 & 25.2 & 57.4 & 54.0 & \cellcolor[HTML]{EFEFEF}\textbf{99.7} & 0.0 \\
 & Learn T10 & 15.2 & 0.2 & 35.0 & 0.4 & 0.1 & 25.4 & 67.7 & 46.1 & 59.0 & \cellcolor[HTML]{EFEFEF}\textbf{99.2} \\
\midrule
\multirow{9}{*}{\rotatebox[origin=c]{-90}{\centering 500}} & Learn first 5 tasks & \cellcolor[HTML]{EFEFEF}\textbf{66.2} & \cellcolor[HTML]{EFEFEF}\textbf{46.0} & \cellcolor[HTML]{EFEFEF}\textbf{56.0} & \cellcolor[HTML]{EFEFEF}\textbf{45.4} & \cellcolor[HTML]{EFEFEF}\textbf{98.2} & 0.0 & 0.0 & 0.0 & 0.0 & 0.0 \\
 & Unlearn T2 & 81.6 & \cellcolor[HTML]{EFEFEF}\textbf{16.3} & 74.0 & 58.1 & 95.3 & 0.0 & 0.0 & 0.0 & 0.0 & 0.0 \\
 & Unlearn T4 & 83.2 & 7.0 & 92.1 & \cellcolor[HTML]{EFEFEF}\textbf{3.9} & 86.9 & 0.0 & 0.0 & 0.0 & 0.0 & 0.0 \\
 & Unlearn T5 & 88.4 & 3.2 & 94.2 & 0.9 & \cellcolor[HTML]{EFEFEF}\textbf{62.6} & 0.0 & 0.0 & 0.0 & 0.0 & 0.0 \\
 & Learn T6 & 78.6 & 2.7 & 70.0 & 0.9 & 3.5 & \cellcolor[HTML]{EFEFEF}\textbf{99.3} & 0.0 & 0.0 & 0.0 & 0.0 \\
 & Learn T7 & 84.2 & 2.4 & 58.8 & 0.1 & 3.2 & 60.0 & \cellcolor[HTML]{EFEFEF}\textbf{98.9} & 0.0 & 0.0 & 0.0 \\
 & Learn T8 & 75.6 & 0.6 & 61.9 & 0.7 & 3.1 & 38.9 & 79.2 & \cellcolor[HTML]{EFEFEF}\textbf{98.6} & 0.0 & 0.0 \\
 & Learn T9 & 43.4 & 0.2 & 52.5 & 0.2 & 3.0 & 59.1 & 75.0 & 80.6 & \cellcolor[HTML]{EFEFEF}\textbf{99.1} & 0.0 \\
 & Learn T10 & 46.2 & 0.5 & 50.6 & 0.2 & 3.0 & 49.4 & 67.7 & 71.0 & 54.7 & \cellcolor[HTML]{EFEFEF}\textbf{100} \\
\midrule
\multirow{9}{*}{\rotatebox[origin=c]{-90}{\centering 5120}} & Learn first 5 tasks & \cellcolor[HTML]{EFEFEF}\textbf{95.2} & \cellcolor[HTML]{EFEFEF}\textbf{99.5} & \cellcolor[HTML]{EFEFEF}\textbf{80.1} & \cellcolor[HTML]{EFEFEF}\textbf{74.1} & \cellcolor[HTML]{EFEFEF}\textbf{92.3} & 0.0 & 0.0 & 0.0 & 0.0 & 0.0 \\
 & Unlearn T2 & 95.8 & \cellcolor[HTML]{EFEFEF}\textbf{27.4} & 85.1 & 92.0 & 94.9 & 0.0 & 0.0 & 0.0 & 0.0 & 0.0 \\
 & Unlearn T4 & 96.0 & 3.7 & 94.4 & \cellcolor[HTML]{EFEFEF}\textbf{34.6} & 95.3 & 0.0 & 0.0 & 0.0 & 0.0 & 0.0 \\
 & Unlearn T5 & 97.2 & 3.7 & 96.5 & 47.2 & \cellcolor[HTML]{EFEFEF}\textbf{71.0} & 0.0 & 0.0 & 0.0 & 0.0 & 0.0 \\
 & Learn T6 & 85.6 & 0.7 & 65.3 & 0.6 & 0.7 & \cellcolor[HTML]{EFEFEF}\textbf{99.7} & 0.0 & 0.0 & 0.0 & 0.0 \\
 & Learn T7 & 88.4 & 0.1 & 30.8 & 0.4 & 2.3 & 66.0 & \cellcolor[HTML]{EFEFEF}\textbf{99.1} & 0.0 & 0.0 & 0.0 \\
 & Learn T8 & 88.4 & 0.1 & 80.9 & 0.2 & 0.2 & 69.0 & 92.3 & \cellcolor[HTML]{EFEFEF}\textbf{98.0} & 0.0 & 0.0 \\
 & Learn T9 & 81.2 & 0.1 & 82.2 & 0.2 & 0.2 & 85.6 & 93.5 & 94.2 & \cellcolor[HTML]{EFEFEF}\textbf{97.9} & 0.0 \\
 & Learn T10 & 78.2 & 0.1 & 69.9 & 0.2 & 0.2 & 83.0 & 92.7 & 91.5 & 86.2 & \cellcolor[HTML]{EFEFEF}\textbf{99.4} \\
\bottomrule
\end{tabular}}
\caption{CL and multiple task UL in CIFAR-10 in a $1\times10$ task distribution setup. UL of Task 2, Task 4, and Task 5 can be observed with accuracy drop to $\sim 0.1\%-70.0\%$ for the corresponding classes. Similarly CL accuracy gains in new Tasks are highlighted with~\textbf{bold}.}
\label{tab:cifar10_1x10}
\vspace{-1\baselineskip}
\end{table}

\begin{table}[t]
    \centering
    \resizebox{\linewidth}{!}{
    \begin{tabular}{c c c c c c c c c c c c}
        \toprule
        \multirow{2}{*}{BS} & Execution & \multicolumn{2}{c}{Task 1} & \multicolumn{2}{c}{Task 2} & \multicolumn{2}{c}{Task 3} & \multicolumn{2}{c}{Task 4} & \multicolumn{2}{c}{Task 5} \\ 
        \cmidrule{3-12}
          &  Sequence& C0 & C1 & C2 & C3 & C4 & C5 & C6 & C7 & C8 & C9 \\ \toprule
        \multirow{6}{*}{\rotatebox[origin=c]{-90}{\centering 200}} & Learn T1 & \cellcolor[HTML]{EFEFEF}\textbf{98.8} & \cellcolor[HTML]{EFEFEF}\textbf{99.5} & 0.0 & 0.0 & 0.0 & 0.0 & 0.0 & 0.0 & 0.0 & 0.0 \\ 
         & Learn T2 & 57.3 & 89.6 & \cellcolor[HTML]{EFEFEF}\textbf{96.3} & \cellcolor[HTML]{EFEFEF}\textbf{97.0} & 0.0 & 0.0 & 0.0 & 0.0 & 0.0 & 0.0 \\ 
         & Unlearn T2 & 99.8 & 94.1 & \cellcolor[HTML]{EFEFEF}\textbf{0.2} & \cellcolor[HTML]{EFEFEF}\textbf{0.3} & 0.0 & 0.0 & 0.0 & 0.0 & 0.0 & 0.0 \\ 
         & Learn T3 & 74.9 & 93.4 & 0.2 & 0.1 & \cellcolor[HTML]{EFEFEF}\textbf{96.9} & \cellcolor[HTML]{EFEFEF}\textbf{95.3} & 0.0 & 0.0 & 0.0 & 0.0 \\ 
         & Learn T4 & 58.6 & 85.9 & 0.2 & 0.3 & 32.1 & 40.9 & \cellcolor[HTML]{EFEFEF}\textbf{99.5} & \cellcolor[HTML]{EFEFEF}\textbf{99.2} & 0.0 & 0.0 \\ 
         & Learn T5 & 34.6 & 58.0 & 0.1 & 0.2 & 51.0 & 60.4 & 81.9 & 69.9 & \cellcolor[HTML]{EFEFEF}\textbf{98.4} & \cellcolor[HTML]{EFEFEF}\textbf{98.8} \\ \midrule
        \multirow{6}{*}{\rotatebox[origin=c]{-90}{\centering 500}} & Learn T1 & \cellcolor[HTML]{EFEFEF}\textbf{99.0} & \cellcolor[HTML]{EFEFEF}\textbf{99.7} & 0.0 & 0.0 & 0.0 & 0.0 & 0.0 & 0.0 & 0.0 & 0.0 \\ 
         & Learn T2 & 74.1 & 96.0 & \cellcolor[HTML]{EFEFEF}\textbf{96.4} & \cellcolor[HTML]{EFEFEF}\textbf{95.3} & 0.0 & 0.0 & 0.0 & 0.0 & 0.0 & 0.0 \\ 
         & Unlearn T2 & 99.6 & 97.1 & \cellcolor[HTML]{EFEFEF}\textbf{0.2} & \cellcolor[HTML]{EFEFEF}\textbf{0.1} & 0.0 & 0.0 & 0.0 & 0.0 & 0.0 & 0.0 \\ 
         & Learn T3 & 75.5 & 96.4 & 0.1 & 0.2 & \cellcolor[HTML]{EFEFEF}\textbf{98.1} & \cellcolor[HTML]{EFEFEF}\textbf{95.7} & 0.0 & 0.0 & 0.0 & 0.0 \\ 
         & Learn T4 & 72.4 & 94.3 & 0.2 & 0.2 & 59.7 & 38.9 & \cellcolor[HTML]{EFEFEF}\textbf{99.7} & \cellcolor[HTML]{EFEFEF}\textbf{97.8} & 0.0 & 0.0 \\ 
         & Learn T5 & 17.0 & 51.9 & 0.2 & 0.2 & 68.9 & 43.8 & 76.1 & 58.2 & \cellcolor[HTML]{EFEFEF}\textbf{98.4} & \cellcolor[HTML]{EFEFEF}\textbf{99.6} \\ \midrule
        \multirow{6}{*}{\rotatebox[origin=c]{-90}{\centering 5120}} & Learn T1 & \cellcolor[HTML]{EFEFEF}\textbf{99.4} & \cellcolor[HTML]{EFEFEF}\textbf{99.2} & 0.0 & 0.0 & 0.0 & 0.0 & 0.0 & 0.0 & 0.0 & 0.0 \\ 
         & Learn T2 & 95.9 & 100.0 & \cellcolor[HTML]{EFEFEF}\textbf{93.5} & \cellcolor[HTML]{EFEFEF}\textbf{94.1} & 0.0 & 0.0 & 0.0 & 0.0 & 0.0 & 0.0 \\ 
         & Unlearn T2 & 98.8 & 100.0 & \cellcolor[HTML]{EFEFEF}\textbf{0.1} & \cellcolor[HTML]{EFEFEF}\textbf{0.3} & 0.0 & 0.0 & 0.0 & 0.0 & 0.0 & 0.0 \\ 
         & Learn T3 & 97.1 & 99.5 & 0.2 & 0.2 & \cellcolor[HTML]{EFEFEF}\textbf{98.1} & \cellcolor[HTML]{EFEFEF}\textbf{94.7} & 0.0 & 0.0 & 0.0 & 0.0 \\ 
         & Learn T4 & 97.7 & 98.3 & 0.4 & 0.1 & 89.0 & 83.1 & \cellcolor[HTML]{EFEFEF}\textbf{98.5} & \cellcolor[HTML]{EFEFEF}\textbf{96.4} & 0.0 & 0.0 \\ 
         & Learn T5 & 89.4 & 88.5 & 0.1 & 0.2 & 89.4 & 84.9 & 95.7 & 89.3 & \cellcolor[HTML]{EFEFEF}\textbf{97.4} & \cellcolor[HTML]{EFEFEF}\textbf{98.0} \\ \bottomrule
    \end{tabular}}
    \caption{CL and single task UL in ciFAIR-10 in a $2\times5$ task distribution. UL of Task 2 can be observed with accuracy dropping to $\sim 0.1\%-0.4\%$ for the corresponding classes. Similarly CL accuracy gains in new Tasks are highlighted with~\textbf{bold}.}
    \label{tab:cifair10_2x5_1}
    \vspace{-1\baselineskip}
\end{table}

\begin{table}[t]
    \centering
    \resizebox{\linewidth}{!}{
    \begin{tabular}{c c c c c c c c c c c c c}
        \toprule
        \multirow{2}{*}{BS} & Execution & \multicolumn{2}{c}{Task 1} & \multicolumn{2}{c}{Task 2} & \multicolumn{2}{c}{Task 3} & \multicolumn{2}{c}{Task 4} & \multicolumn{2}{c}{Task 5} \\ \cmidrule{3-12}
         & Sequence& C0 & C1 & C2 & C3 & C4 & C5 & C6 & C7 & C8 & C9 \\ \toprule
        \multirow{4}{*}{\rotatebox[origin=c]{-90}{\centering 200}} & Learn initial 4 tasks & \cellcolor[HTML]{EFEFEF}\textbf{67.8} & \cellcolor[HTML]{EFEFEF}\textbf{93.9} & \cellcolor[HTML]{EFEFEF}\textbf{56.5} & \cellcolor[HTML]{EFEFEF}\textbf{31.1} & \cellcolor[HTML]{EFEFEF}\textbf{58.7} & \cellcolor[HTML]{EFEFEF}\textbf{58.8} & \cellcolor[HTML]{EFEFEF}\textbf{99.5} & \cellcolor[HTML]{EFEFEF}\textbf{96.4} & 0.0 & 0.0 \\ 
         & Unlearn T2 & 78.2 & 97.6 & \cellcolor[HTML]{EFEFEF}\textbf{1.1} & \cellcolor[HTML]{EFEFEF}\textbf{0.5} & 97.7 & 83.1 & 98.0 & 89.9 & 0.0 & 0.0 \\ 
         & Unlearn T4 & 62.7 & 94.6 & 0.01 & 0.1 & 93.9 & 90.2 & \cellcolor[HTML]{EFEFEF}\textbf{0.2} & \cellcolor[HTML]{EFEFEF}\textbf{0.3} & 0.0 & 0.0 \\ 
         & Learn T5 & 19.6 & 40.5 & 0.1 & 0.1 & 86.5 & 77.8 & 0.2 & 0.1 & \cellcolor[HTML]{EFEFEF}\textbf{97.4} & \cellcolor[HTML]{EFEFEF}\textbf{99.4} \\ \midrule
        \multirow{4}{*}{\rotatebox[origin=c]{-90}{\centering 500}} & Learn initial 4 tasks & \cellcolor[HTML]{EFEFEF}\textbf{76.0} & \cellcolor[HTML]{EFEFEF}\textbf{93.2} & \cellcolor[HTML]{EFEFEF}\textbf{68.3} & \cellcolor[HTML]{EFEFEF}\textbf{37.3} & \cellcolor[HTML]{EFEFEF}\textbf{80.5} & \cellcolor[HTML]{EFEFEF}\textbf{68.5} & \cellcolor[HTML]{EFEFEF}\textbf{98.0} & \cellcolor[HTML]{EFEFEF}\textbf{98.8} & 0.0 & 0.0 \\ 
         & Unlearn T2 & 93.6 & 96.4 & \cellcolor[HTML]{EFEFEF}\textbf{4.0} & \cellcolor[HTML]{EFEFEF}\textbf{2.0} & 96.1 & 89.0 & 99.3 & 90.0 & 0.0 & 0.0 \\ 
         & Unlearn T4 & 87.0 & 94.6 & 2.2 & 0.2 & 97.9 & 95.1 & \cellcolor[HTML]{EFEFEF}\textbf{0.2} & \cellcolor[HTML]{EFEFEF}\textbf{0.3} & 0.0 & 0.0 \\ 
         & Learn T5 & 43.7 & 70.9 & 0.2 & 0.3 & 90.6 & 88.8 & 0.1 & 0.1 & \cellcolor[HTML]{EFEFEF}\textbf{80.0} & \cellcolor[HTML]{EFEFEF}\textbf{99.0} \\ \midrule
        \multirow{4}{*}{\rotatebox[origin=c]{-90}{\centering 5120}} & Learn initial 4 tasks & \cellcolor[HTML]{EFEFEF}\textbf{96.9} & \cellcolor[HTML]{EFEFEF}\textbf{99.2} & \cellcolor[HTML]{EFEFEF}\textbf{88.9} & \cellcolor[HTML]{EFEFEF}\textbf{79.9} & \cellcolor[HTML]{EFEFEF}\textbf{87.8} & \cellcolor[HTML]{EFEFEF}\textbf{80.3} & \cellcolor[HTML]{EFEFEF}\textbf{96.4} & \cellcolor[HTML]{EFEFEF}\textbf{96.4} & 0.0 & 0.0 \\ 
         & Unlearn T2 & 98.8 & 99.2 & \cellcolor[HTML]{EFEFEF}\textbf{46.2} & \cellcolor[HTML]{EFEFEF}\textbf{19.8} & 96.1 & 90.8 & 86.6 & 97.4 & 0.0 & 0.0 \\ 
         & Unlearn T4 & 97.1 & 99.5 & 36.2 & 3.8 & 98.5 & 94.9 & \cellcolor[HTML]{EFEFEF}\textbf{1.4} & \cellcolor[HTML]{EFEFEF}\textbf{0.7} & 0.0 & 0.0 \\ 
         & Learn T5 & 92.1 & 94.6 & 0.1 & 0.2 & 92.8 & 93.1 & 0.2 & 2.3 & \cellcolor[HTML]{EFEFEF}\textbf{93.4} & \cellcolor[HTML]{EFEFEF}\textbf{91.2} \\ \bottomrule
    \end{tabular}}
    \caption{CL and multiple task UL in ciFAIR-10 in a $2\times5$ task distribution. UL of Task 2 and Task 4 can be observed with accuracy dropping to $\sim 0.1\%-46\%$ for the corresponding classes. Similarly CL accuracy gains in new Tasks are highlighted with~\textbf{bold}.}
    \label{tab:cifair10_2x5_4}
    \vspace{-1\baselineskip}
\end{table}

\begin{table}[t]
\centering
\resizebox{\linewidth}{!}{
\begin{tabular}{ l  l  l  l  l  l  l  l  l  l  l  l}
\toprule
\multirow{2}{*}{BS} & Execution & T1 & T2 & T3 & T4 & T5 & T6 & T7 & T8 & T9 & T10 \\ \cmidrule{3-12}
 &  Sequence & C0 & C1 & C2 & C3 & C4 & C5 & C6 & C7 & C8 & C9 \\
\toprule
\multirow{9}{*}{\rotatebox[origin=c]{-90}{\centering 200}} & Learn first 5 tasks & \cellcolor[HTML]{EFEFEF}\textbf{53.3} & \cellcolor[HTML]{EFEFEF}\textbf{90.6} & \cellcolor[HTML]{EFEFEF}\textbf{30.3} & \cellcolor[HTML]{EFEFEF}\textbf{24.3} & \cellcolor[HTML]{EFEFEF}\textbf{98.7} & 0.0 & 0.0 & 0.0 & 0.0 & 0.0 \\
 & Unlearn T2 & 49.7 & \cellcolor[HTML]{EFEFEF}\textbf{0.2} & 57.8 & 53.0 & 95.3 & 0.0 & 0.0 & 0.0 & 0.0 & 0.0 \\
 & Unlearn T4 & 63.2 & 0.2 & 94.6 & \cellcolor[HTML]{EFEFEF}\textbf{0.3} & 78.2 & 0.0 & 0.0 & 0.0 & 0.0 & 0.0 \\
 & Unlearn T5 & 56.9 & 0.2 & 99.4 & 0.3 & \cellcolor[HTML]{EFEFEF}\textbf{25.0} & 0.0 & 0.0 & 0.0 & 0.0 & 0.0 \\
 & Learn T6 & 47.8 & 0.1 & 49.5 & 0.2 & 0.2 & \cellcolor[HTML]{EFEFEF}\textbf{99.7} & 0.0 & 0.0 & 0.0 & 0.0 \\
 & Learn T7 & 65.7 & 0.2 & 55.0 & 0.2 & 0.2 & 46.2 & \cellcolor[HTML]{EFEFEF}\textbf{98.5} & 0.0 & 0.0 & 0.0 \\
 & Learn T8 & 64.8 & 0.1 & 51.3 & 0.2 & 0.1 & 26.1 & 63.4 & \cellcolor[HTML]{EFEFEF}\textbf{99.0} & 0.0 & 0.0 \\
 & Learn T9 & 8.9 & 0.1 & 24.0 & 0.2 & 0.2 & 22.7 & 51.4 & 16.0 & \cellcolor[HTML]{EFEFEF}\textbf{100} & 0.0 \\
 & Learn T10 & 30.0 & 0.1 & 26.8 & 0.1 & 0.2 & 17.4 & 45.1 & 12.0 & 33.3 & \cellcolor[HTML]{EFEFEF}\textbf{100} \\
\midrule
\multirow{9}{*}{\rotatebox[origin=c]{-90}{\centering 500}} & Learn first 5 tasks & \cellcolor[HTML]{EFEFEF}\textbf{66.3} & \cellcolor[HTML]{EFEFEF}\textbf{90.1} & \cellcolor[HTML]{EFEFEF}\textbf{38.8} & \cellcolor[HTML]{EFEFEF}\textbf{46.5} & \cellcolor[HTML]{EFEFEF}\textbf{97.5} & 0.0 & 0.0 & 0.0 & 0.0 & 0.0 \\
 & Unlearn T2 & 64.8 & \cellcolor[HTML]{EFEFEF}\textbf{2.3} & 66.1 & 64.8 & 96.3 & 0.0 & 0.0 & 0.0 & 0.0 & 0.0 \\
 & Unlearn T4 & 76.8 & 0.2 & 93.7 & \cellcolor[HTML]{EFEFEF}\textbf{1.4} & 81.0 & 0.0 & 0.0 & 0.0 & 0.0 & 0.0 \\
 & Unlearn T5 & 74.9 & 0.4 & 97.0 & 0.1 & \cellcolor[HTML]{EFEFEF}\textbf{43.0} & 0.0 & 0.0 & 0.0 & 0.0 & 0.0 \\
 & Learn T6 & 16.2 & 0.4 & 17.9 & 0.2 & 0.2 & \cellcolor[HTML]{EFEFEF}\textbf{99.7} & 0.0 & 0.0 & 0.0 & 0.0 \\
 & Learn T7 & 4.0 & 0.2 & 26.6 & 0.1 & 0.3 & 43.6 & \cellcolor[HTML]{EFEFEF}\textbf{99.7} & 0.0 & 0.0 & 0.0 \\
 & Learn T8 & 1.3 & 0.1 & 40.2 & 0.1 & 0.2 & 28.3 & 72.5 & \cellcolor[HTML]{EFEFEF}\textbf{99.4} & 0.0 & 0.0 \\
 & Learn T9 & 2.7 & 0.1 & 13.8 & 0.2 & 0.2 & 21.9 & 34.4 & 58.6 & \cellcolor[HTML]{EFEFEF}\textbf{100} & 0.0 \\
 & Learn T10 & 2.3 & 0.1 & 34.5 & 0.2 & 0.2 & 20.2 & 47.4 & 56.2 & 48.6 & \cellcolor[HTML]{EFEFEF}\textbf{100} \\
\midrule
\multirow{9}{*}{\rotatebox[origin=c]{-90}{\centering 5120}} & Learn first 5 tasks & \cellcolor[HTML]{EFEFEF}\textbf{86.0} & \cellcolor[HTML]{EFEFEF}\textbf{99.5} & \cellcolor[HTML]{EFEFEF}\textbf{77.4} & \cellcolor[HTML]{EFEFEF}\textbf{79.3} & \cellcolor[HTML]{EFEFEF}\textbf{91.8} & 0.0 & 0.0 & 0.0 & 0.0 & 0.0 \\
 & Unlearn T2 & 96.1 & \cellcolor[HTML]{EFEFEF}\textbf{25.2} & 90.5 & 82.8 & 88.6 & 0.0 & 0.0 & 0.0 & 0.0 & 0.0 \\
 & Unlearn T4 & 95.7 & 10.5 & 96.6 & \cellcolor[HTML]{EFEFEF}\textbf{58.3} & 88.8 & 0.0 & 0.0 & 0.0 & 0.0 & 0.0 \\
 & Unlearn T5 & 95.9 & 6.3 & 97.4 & 60.5 & \cellcolor[HTML]{EFEFEF}\textbf{64.2} & 0.0 & 0.0 & 0.0 & 0.0 & 0.0 \\
 & Learn T6 & 86.9 & 0.2 & 41.0 & 0.3 & 0.3 & \cellcolor[HTML]{EFEFEF}\textbf{100} & 0.0 & 0.0 & 0.0 & 0.0 \\
 & Learn T7 & 57.5 & 0.2 & 83.1 & 0.1 & 0.1 & 79.1 &\cellcolor[HTML]{EFEFEF}\textbf{ 98.9} & 0.0 & 0.0 & 0.0 \\
 & Learn T8 & 61.5 & 0.2 & 82.0 & 0.2 & 0.1 & 61.2 & 93.1 & \cellcolor[HTML]{EFEFEF}\textbf{98.0} & 0.0 & 0.0 \\
 & Learn T9 & 67.4 & 0.2 & 83.3 & 0.2 & 0.1 & 79.9 & 92.7 & 90.2 & \cellcolor[HTML]{EFEFEF}\textbf{97.6} & 0.0 \\
 & Learn T10 & 84.1 & 0.2 & 86.8 & 0.1 & 0.1 & 78.2 & 95.9 & 82.3 & 90.1 & \cellcolor[HTML]{EFEFEF}\textbf{98.8} \\
\bottomrule
\end{tabular}}
\caption{CL and multiple task UL in ciFAIR-10 in a $1\times10$ task distribution setup. UL of Task 2, Task 4, and Task 5 can be observed with accuracy dropped to $\sim 0.1\%-64\%$ for the corresponding classes. Similarly CL accuracy gains in new Tasks are highlighted with~\textbf{bold}.}
\label{tab:cifair10_1x10}
\vspace{-1\baselineskip}
\end{table}

\section{Experiments and Results}
\subsection{Experiment Setting}
\textbf{Datasets and experimental settings.} We conducted our experiments using the CIFAR-10 and ciFAIR-10 datasets with NVIDIA A6000 48 GB GPU. Buffer sizes of 200, 500, and 5120 were tested, with Bernoulli probability and ER weight set to 0.2 and 0.5 for smaller buffers, and 0.8 and 1.0 for the larger buffer. This setup was consistent across all experiments. Our base model was ResNet-18. Additionally, our implementation added two projector heads: an equivariance head and an invariance head, each with three linear layers, batch normalization, and ReLU activations. The equivariance head outputs 4 dimensions, while the invariance head outputs 64 dimensions, with the latter being normalized.\par

\textbf{Baselines.} Our method is the first to introduce a unified framework that addresses both unlearning (UL) and continual learning (CL), bridging the gap between these two traditionally distinct fields. In the absence of existing approaches that integrate UL and CL, the most appropriate baselines for comparison are the independent results obtained by state-of-the-art methods in each field. These baselines enable a rigorous evaluation of our framework's effectiveness in tackling tasks typically handled separately by UL and CL, establishing a comprehensive benchmark for performance assessment.

\subsection{Single task CL-UL with $2\times5$ distribution} 
The impact of varying buffer sizes on continual learning and single task unlearning in a $2\times5$ task distribution setting is illustrated in Table~\ref{tab:cifar10_2x5_1} for CIFAR-10. For a buffer size of 200, unlearning Task 2 results in $\sim0 \%$ accuracy for that task, indicating effective unlearning. However, this process also leads to a significant drop in the accuracy of Task 1, with a decrease from $99.4\%$ to $87.4\%$ reflecting the challenge of retaining prior knowledge. As the buffer size increases to 500, the retention of Task 1 accuracy after unlearning Task 2 improves, dropping only to $92.1\%$. Similarly, other tasks show stable performance with minimal accuracy loss. When the buffer size is further increased to 5120, the retention of accuracy improves significantly across all tasks, with Task 1 maintaining $97.2\%$ accuracy after unlearning Task 2. This trend highlights that larger buffer sizes are more effective in preserving task accuracy during the unlearning process.\par
Results on ciFAIR-10 (Table~\ref{tab:cifair10_2x5_1}) shows effective unlearning of Task 2 across all buffer sizes, as evidenced by the post-unlearning accuracy dropping to $\sim0\%$. However, the retained accuracy for Task 1 exhibits a dip if the buffer size is increased from 200 to 500, showing an average drop of $52.9\%$ and $64.9\%$ respectively. In contrast, a larger buffer size of 5120 results in better retention, with a noticeable improvement across all tasks, especially in the retention of Task 1.\par

\subsection{Multiple task CL-UL with $2\times5$ distribution} 
The impact of varying buffer sizes on continual learning and multiple task unlearning in a $2\times5$ task distribution setting is illustrated in Table~\ref{tab:cifar10_2x5_4} for CIFAR-10. Its showcases similar trends of performance retention in previously learned tasks with respect to the buffer size. With a buffer size of 200, unlearning both Task 2 and Task 4 leads to $0\%$ accuracy on these tasks. Interestingly, the poor performance on the classes in Task 3 is recovered to a large extent (C4:$53.3\%\rightarrow 97.6\%$, C5:$65.2\%\rightarrow 94.5\%$) after unkearning these tasks. However, learning task 5 leads to a signifcant drop in task 1's performance (C0:$70.0\%\rightarrow 19.6\%$, C1:$79.5\%\rightarrow 39.4\%$). Increasing the buffer size to 500 mitigates this effect to certain extent, with the classes in Task 1 retaining $47.8\%$ and $65.9\%$ accuracy. With the largest buffer size of 5120, the model exhibits excellent accuracy retention across tasks, with the model retainig $84.8\%$ percent after all the learning and unlearning requests.\par
Results in ciFAIR-10 (Table~\ref{tab:cifair10_2x5_4}) shows that unlearning Task 2 and Task 4 results in successful information removal for all buffer sizes, achieving an accuracy of $\approx 0\%$ post-unlearning. The increase in buffer size results in better retention accuracy across the retain classes, with significant retention for Task 1 in buffer size 5120. With a buffer size of 5120, the model only loses an average $4.7\%$ of the accuracy for Task 1 as compared to an average deterioration of $50.8\%$ with a buffer size of 200.\par

\subsection{Multiple task CL-UL with $1\times10$ distribution}
The impact of varying buffer sizes on continual learning and multiple task unlearning in a $1\times10$ task distribution setting is illustrated in Table~\ref{tab:cifar10_1x10} for CIFAR-10 which expands the complexity by testing multiple task unlearning and yet shows similar trends as Table \ref{tab:cifar10_2x5_1} and Table \ref{tab:cifar10_2x5_4}. For a buffer size of 200, the results show substantial accuracy losses, particularly for Task 1, where accuracy drops from $37.2\%$ to $9.6\%$ after the eight request to learn task 9. When the buffer size is increased to 500, the retention of Task 1’s accuracy improves, with accuracy dropping to $46.2\%$ after all CL-UL requests. Nevertheless, unlearning continues to significantly affect the retention of information across other tasks. With a buffer size of 5120, the model demonstrates strong performance, retaining $78.2\%$ accuracy for Task 1 even after extensive learning and unlearning operations.\par
In ciFAIR-10 (Table~\ref{tab:cifair10_1x10}), the unlearning of Task 2, Task 4, and Task 5 leads to successful information removal, achieving $0.0\%$ accuracy for these tasks in all buffer sizes. Similar to previously discussed results, an increase in buffer size correlates with better retention of task accuracies. Specifically, buffer size 5120 demonstrates the most robust retention across all tasks, with minimal accuracy loss (only $1.9\%$ for task 1) for retained classes. In contrast, buffer sizes 200 and 500 show significant drops in the retention accuracy of Task 1, with a performance drop of $23.3\%$ and $66\%$ respectively. The pattern observed suggests that larger buffer sizes are crucial in maintaining the integrity of retained tasks, particularly in more complex learning-unlearning sequences involving a higher number of tasks.\par

\begin{table}[t]
\centering
\resizebox{\linewidth}{!}{%
\begin{tabular}{l l  l  l  l  l  l  l  l  l  l  l}
\toprule
\multirow{2}{*}{BS} & Execution & \multicolumn{2}{l}{Task 1} & \multicolumn{2}{l}{Task2} & \multicolumn{2}{l}{Task 3} & \multicolumn{2}{l}{Task 4} & \multicolumn{2}{l}{Task 5} \\  
\cmidrule{3-12}
 &  Sequence& C0 & C1 & C2 & C3 & C4 & C5 & C6 & C7 & C8 & C9 \\
\toprule
\multirow{2}{*}{200} & Baseline(retrain) & 44.6 & 53.0 & \cellcolor[HTML]{EFEFEF}\textbf{0.2} & \cellcolor[HTML]{EFEFEF}\textbf{0.3} & 87.8 & 81.0 & \cellcolor[HTML]{EFEFEF}\textbf{0.3} & \cellcolor[HTML]{EFEFEF}\textbf{0.1} & 98.9 & 99.6 \\ 
 & Our method & 19.6 & 39.4 & \cellcolor[HTML]{EFEFEF}\textbf{0.1} & \cellcolor[HTML]{EFEFEF}\textbf{0.3} & 81.8 & 83.4 & \cellcolor[HTML]{EFEFEF}\textbf{0.3} & \cellcolor[HTML]{EFEFEF}\textbf{0.1} & 98.1 & 99.8 \\
\midrule
\multirow{2}{*}{500} & Baseline(retrain) & 72.0 & 67.5 & \cellcolor[HTML]{EFEFEF}\textbf{0.2} & \cellcolor[HTML]{EFEFEF}\textbf{0.2} & 90.8 & 83.0 & \cellcolor[HTML]{EFEFEF}\textbf{0.1} & \cellcolor[HTML]{EFEFEF}\textbf{0.2} & 97.5 & 99.4 \\ 
 & Our method & 47.8 & 65.9 & \cellcolor[HTML]{EFEFEF}\textbf{0.1} & \cellcolor[HTML]{EFEFEF}\textbf{0.01} & 90.2 & 88.0 & \cellcolor[HTML]{EFEFEF}\textbf{0.1} & \cellcolor[HTML]{EFEFEF}\textbf{0.2} & 99.1 & 99.2 \\
\midrule
\multirow{2}{*}{5120} & Baseline(retrain) & 93.8 & 94.6 & \cellcolor[HTML]{EFEFEF}\textbf{0.1} & \cellcolor[HTML]{EFEFEF}\textbf{0.4} & 96.6 & 90.7 & \cellcolor[HTML]{EFEFEF}\textbf{0.2} & \cellcolor[HTML]{EFEFEF}\textbf{0.02} & 95.2 & 97.2 \\
 & Our method & 84.8 & 90.2 & \cellcolor[HTML]{EFEFEF}\textbf{0.1} & \cellcolor[HTML]{EFEFEF}\textbf{0.1} & 96.0 & 95.1 & \cellcolor[HTML]{EFEFEF}\textbf{0.25} & \cellcolor[HTML]{EFEFEF}\textbf{0.2} & 98.7 & 98.6 \\
\bottomrule 
\end{tabular}}
\caption{Comparison of the proposed method with a \textit{baseline UL (retrained) model} on CIFAR-10.}
\label{tab:ul_comparison}
\end{table}

\begin{table}[t]
\centering
\resizebox{\linewidth}{!}{%
\begin{tabular}{l l l l l l l l l}
\toprule
\multirow{4}{*}{BS} & \multirow{4}{*}{DER} & \multicolumn{2}{l}{oEWC (B.A.)} & \multicolumn{2}{l}{SGD(B.A.)} & \multicolumn{2}{l}{Our Method} \\  
\cmidrule{3-8}
 &  & All & Retain & All & Retain & Before UL & After UL \\
 &&classes&classes&classes&classes&(retain&(retain\\
 && &&& &classes)&classes)\\
\toprule
200 & 61.93 & \multirow{3}{*}{19.25} & \multirow{3}{*}{32.35} & \multirow{3}{*}{19.45} & \multirow{3}{*}{30.91} & \cellcolor[HTML]{EFEFEF}\textbf{65.7} & \cellcolor[HTML]{EFEFEF}\textbf{70.35} \\ 
500 & 70.51 &  &  &  &  & \cellcolor[HTML]{EFEFEF}\textbf{76.9} & \cellcolor[HTML]{EFEFEF}\textbf{81.7} \\ 
5120 & 83.81 &  &  &  &  & \cellcolor[HTML]{EFEFEF}\textbf{89.5} & \cellcolor[HTML]{EFEFEF}\textbf{93.9} \\
\bottomrule 
\end{tabular}}
\caption{Comparison of the proposed method with  baseline CL methods on CIFAR-10. Buffer based: DER~\cite{buzzega2020dark}, Buffer Agnostic (B.A.): oEWC~\cite{schwarz2018progress}, SGD.}
\vspace{-1\baselineskip}
\label{tab:cl_comparison}
\end{table}

\begin{table}[t]
\centering
\resizebox{\linewidth}{!}{%
\begin{tabular}{ l l l l l  l  l  l  l  l  l  l}
\toprule
\multirow{2}{*}{BS} & Execution & \multicolumn{5}{l}{Task 1} & \multicolumn{5}{l}{Task 2} \\
\cmidrule{3-12}
 & Sequence & C0 & C1 & C2 & C3 & C4 & C5 & C6 & C7 & C8 & C9 \\
\toprule
\multirow{2}{*}{\rotatebox[origin=c]{-90}{\centering 200}} & Learn 2 Tasks & \cellcolor[HTML]{EFEFEF}\textbf{54.0} & \cellcolor[HTML]{EFEFEF}\textbf{55.5} & \cellcolor[HTML]{EFEFEF}\textbf{58.4} & \cellcolor[HTML]{EFEFEF}\textbf{22.5} & \cellcolor[HTML]{EFEFEF}\textbf{52.1} & \cellcolor[HTML]{EFEFEF}\textbf{93.1} & \cellcolor[HTML]{EFEFEF}\textbf{98.0} & \cellcolor[HTML]{EFEFEF}\textbf{97.4} & \cellcolor[HTML]{EFEFEF}\textbf{97.9} & \cellcolor[HTML]{EFEFEF}\textbf{99.8} \\
 & Unlearn T2 & 76.8 & 94.0 & 74.0 & 81.8 & 94.1 & \cellcolor[HTML]{EFEFEF}\textbf{0.4} & \cellcolor[HTML]{EFEFEF}\textbf{0.6} & \cellcolor[HTML]{EFEFEF}\textbf{0.45} & \cellcolor[HTML]{EFEFEF}\textbf{0.4} & \cellcolor[HTML]{EFEFEF}\textbf{0.53} \\
\midrule
\multirow{2}{*}{\rotatebox[origin=c]{-90}{\centering 500}} & Learn 2 Tasks & \cellcolor[HTML]{EFEFEF}\textbf{77.6} & \cellcolor[HTML]{EFEFEF}\textbf{84.7} & \cellcolor[HTML]{EFEFEF}\textbf{74.4} & \cellcolor[HTML]{EFEFEF}\textbf{66.1} & \cellcolor[HTML]{EFEFEF}\textbf{65.8} & \cellcolor[HTML]{EFEFEF}\textbf{85.0} & \cellcolor[HTML]{EFEFEF}\textbf{97.8} & \cellcolor[HTML]{EFEFEF}\textbf{97.4} & \cellcolor[HTML]{EFEFEF}\textbf{97.9} & \cellcolor[HTML]{EFEFEF}\textbf{98.8} \\
 & Unlearn T2 & 95.4 & 97.5 & 78.0 & 96.2 & 75.7 & \cellcolor[HTML]{EFEFEF}\textbf{5.4} & \cellcolor[HTML]{EFEFEF}\textbf{37.8} & \cellcolor[HTML]{EFEFEF}\textbf{52.8} & \cellcolor[HTML]{EFEFEF}\textbf{28.0} & \cellcolor[HTML]{EFEFEF}\textbf{20.0} \\
\midrule
\multirow{2}{*}{\rotatebox[origin=c]{-90}{\centering 5120}} & Learn 2 Tasks & \cellcolor[HTML]{EFEFEF}\textbf{96.0} & \cellcolor[HTML]{EFEFEF}\textbf{95.0} & \cellcolor[HTML]{EFEFEF}\textbf{88.5} & \cellcolor[HTML]{EFEFEF}\textbf{79.6} & \cellcolor[HTML]{EFEFEF}\textbf{91.9} & \cellcolor[HTML]{EFEFEF}\textbf{80.4} & \cellcolor[HTML]{EFEFEF}\textbf{95.7} & \cellcolor[HTML]{EFEFEF}\textbf{94.4} & \cellcolor[HTML]{EFEFEF}\textbf{94.2} & \cellcolor[HTML]{EFEFEF}\textbf{96.4} \\
 & Unlearn T2 & 97.0 & 97.7 & 93.5 & 89.2 & 96.4 & \cellcolor[HTML]{EFEFEF}\textbf{26.4} & \cellcolor[HTML]{EFEFEF}\textbf{11.9} & \cellcolor[HTML]{EFEFEF}\textbf{7.9} & \cellcolor[HTML]{EFEFEF}\textbf{22.7} & \cellcolor[HTML]{EFEFEF}\textbf{23.8} \\ \bottomrule
\end{tabular}}
\caption{CL and single task UL in CIFAR-10 in a $5\times2$ task distribution. UL of Task 2 can be observed with accuracy dropping to $\sim 0.4\%-25\%$ for corresponding classes.}
\label{tab:cifar10_5x2}
\vspace{-1\baselineskip}
\end{table}

\begin{table}[t]
\centering
\resizebox{\linewidth}{!}{%
\begin{tabular}{l  l  l  l  l  l  l  l  l  l  l  l}
\toprule
\multirow{2}{*}{BP} & Execution & \multicolumn{2}{l}{Task 1} & \multicolumn{2}{l}{Task 2} & \multicolumn{2}{l}{Task 3} & \multicolumn{2}{l}{Task 4} & \multicolumn{2}{l}{Task 5} \\
\cmidrule{3-12}
 &  Sequence& C0 & C1 & C2 & C3 & C4 & C5 & C6 & C7 & C8 & C9\\
\toprule
\multirow{2}{*}{\textbf{0.5}} & Learn first 4 tasks & \cellcolor[HTML]{EFEFEF}\textbf{93.6} & \cellcolor[HTML]{EFEFEF}\textbf{98.9} & \cellcolor[HTML]{EFEFEF}\textbf{83.6} & \cellcolor[HTML]{EFEFEF}\textbf{77.2} & \cellcolor[HTML]{EFEFEF}\textbf{88.4} & \cellcolor[HTML]{EFEFEF}\textbf{81.2} & \cellcolor[HTML]{EFEFEF}\textbf{96.3} & \cellcolor[HTML]{EFEFEF}\textbf{95.3} & 0.0 & 0.0 \\
 & End of sequence & \cellcolor[HTML]{EFEFEF}\textbf{84.8} & \cellcolor[HTML]{EFEFEF}\textbf{92.1} & \cellcolor[HTML]{EFEFEF}\textbf{0.2} & \cellcolor[HTML]{EFEFEF}\textbf{0.1} & \cellcolor[HTML]{EFEFEF}\textbf{91.4} & \cellcolor[HTML]{EFEFEF}\textbf{91.3} & \cellcolor[HTML]{EFEFEF}\textbf{0.1} & \cellcolor[HTML]{EFEFEF}\textbf{0.3} & \cellcolor[HTML]{EFEFEF}\textbf{98.3} & \cellcolor[HTML]{EFEFEF}\textbf{97.4} \\
\midrule
\multirow{2}{*}{\textbf{0.6}} & Learn first 4 tasks & \cellcolor[HTML]{EFEFEF}\textbf{95.4} & \cellcolor[HTML]{EFEFEF}\textbf{99.5} & \cellcolor[HTML]{EFEFEF}\textbf{87.4} & \cellcolor[HTML]{EFEFEF}\textbf{70.5} & \cellcolor[HTML]{EFEFEF}\textbf{88.4} & \cellcolor[HTML]{EFEFEF}\textbf{82.6} & \cellcolor[HTML]{EFEFEF}\textbf{94.2} & \cellcolor[HTML]{EFEFEF}\textbf{94.7} & 0.0 & 0.0 \\
 & End of sequence & \cellcolor[HTML]{EFEFEF}\textbf{91.4} & \cellcolor[HTML]{EFEFEF}\textbf{92.9} & \cellcolor[HTML]{EFEFEF}\textbf{0.3} & \cellcolor[HTML]{EFEFEF}\textbf{0.2} & \cellcolor[HTML]{EFEFEF}\textbf{97.2} & \cellcolor[HTML]{EFEFEF}\textbf{94.5} & \cellcolor[HTML]{EFEFEF}\textbf{0.1} & \cellcolor[HTML]{EFEFEF}\textbf{0.2} & \cellcolor[HTML]{EFEFEF}\textbf{97.1} & \cellcolor[HTML]{EFEFEF}\textbf{96.8} \\
\midrule
\multirow{2}{*}{\textbf{0.7}} & Learn first 4 tasks & \cellcolor[HTML]{EFEFEF}\textbf{94.4} & \cellcolor[HTML]{EFEFEF}\textbf{99.7} & \cellcolor[HTML]{EFEFEF}\textbf{80.0} & \cellcolor[HTML]{EFEFEF}\textbf{74.7} & \cellcolor[HTML]{EFEFEF}\textbf{89.6} & \cellcolor[HTML]{EFEFEF}\textbf{84.2} & \cellcolor[HTML]{EFEFEF}\textbf{95.2} & \cellcolor[HTML]{EFEFEF}\textbf{95.9} & 0.0 & 0.0 \\
 & End of sequence & \cellcolor[HTML]{EFEFEF}\textbf{79.8} & \cellcolor[HTML]{EFEFEF}\textbf{86.7} & \cellcolor[HTML]{EFEFEF}\textbf{0.1} & \cellcolor[HTML]{EFEFEF}\textbf{0.2} & \cellcolor[HTML]{EFEFEF}\textbf{96.6} & \cellcolor[HTML]{EFEFEF}\textbf{94.7} & \cellcolor[HTML]{EFEFEF}\textbf{0.01} & \cellcolor[HTML]{EFEFEF}\textbf{0.2} & \cellcolor[HTML]{EFEFEF}\textbf{98.3} & \cellcolor[HTML]{EFEFEF}\textbf{98.6} \\ \bottomrule
\end{tabular}}
\caption{Bernoulli Probability (BP) variations on CIFAR-10 with buffer size 5120. The CL sequence is learning first four tasks then UL Task 2 and Task 4 followed by CL Task 5.}
\label{tab:bp_ablation}
\vspace{-1\baselineskip}
\end{table}

\begin{table}[t]
\centering
\resizebox{\linewidth}{!}{
\begin{tabular}{ l  l  l  l  l  l  l  l  l  l  l  l}
\toprule
\multirow{2}{*}{Loss Removal} & Execution & \multicolumn{2}{c}{Task 1} & \multicolumn{2}{c}{Task 2} & \multicolumn{2}{c}{Task 3} & \multicolumn{2}{c}{Task 4} & \multicolumn{2}{c}{Task 5} \\ \cmidrule{3-12}
 & Sequence & C-0 & C-1 & C-2 & C-3 & C-4 & C-5 & C-6 & C-7 & C-8 & C-9 \\
\toprule
 Online& Learn the first 4 tasks & \cellcolor[HTML]{EFEFEF}\textbf{72.0} & \cellcolor[HTML]{EFEFEF}\textbf{82.4} & \cellcolor[HTML]{EFEFEF}\textbf{56.5} & \cellcolor[HTML]{EFEFEF}\textbf{26.4} & \cellcolor[HTML]{EFEFEF}\textbf{64.8} & \cellcolor[HTML]{EFEFEF}\textbf{41.4} & \cellcolor[HTML]{EFEFEF}\textbf{98.9} & \cellcolor[HTML]{EFEFEF}\textbf{99.0} & 0.0 & 0.0 \\
Distillation & Unlearn T2 & 94 & 91.7 & \cellcolor[HTML]{EFEFEF}\textbf{1.5} & \cellcolor[HTML]{EFEFEF}\textbf{0.2} & 85.3 & 35.1 & 95.5 & 80.6 & 0.0 & 0.0 \\
($\mathcal{L}_{od}$) & Unlearn T4 & 87.4 & 91.9 & 0.9 & 0.1 & 99.6 & 87.6 & \cellcolor[HTML]{EFEFEF}\textbf{26.2} & \cellcolor[HTML]{EFEFEF}\textbf{1.7} & 0.0 & 0.0 \\
 & Learn T5 & 29.4 & 55.1 & 0.2 & 0.2 & 76.9 & 67.4 & 0.02 & 0.1 & \cellcolor[HTML]{EFEFEF}\textbf{98.1} & \cellcolor[HTML]{EFEFEF}\textbf{99.2} \\
\midrule
Contrastive 
 & Learn the first 4 tasks & \cellcolor[HTML]{EFEFEF}\textbf{78.4} & \cellcolor[HTML]{EFEFEF}\textbf{95.0} & \cellcolor[HTML]{EFEFEF}\textbf{71.8} & \cellcolor[HTML]{EFEFEF}\textbf{54.3} & \cellcolor[HTML]{EFEFEF}\textbf{74.4} & \cellcolor[HTML]{EFEFEF}\textbf{69.2} & \cellcolor[HTML]{EFEFEF}\textbf{97.2} & \cellcolor[HTML]{EFEFEF}\textbf{95.9} & 0.0 & 0.0 \\
Distillation  & Unlearn T2 & 86.2 & 96.4 & \cellcolor[HTML]{EFEFEF}\textbf{2.6} & \cellcolor[HTML]{EFEFEF}\textbf{0.1} & 83.0 & 85.0 & 96.5 & 93.0 & 0.0 & 0.0 \\
($\mathcal{L}_{cd})$ & Unlearn T4 & 86.4 & 93.1 & 1.0 & 0.4 & 99.0 & 95.1 & \cellcolor[HTML]{EFEFEF}\textbf{42.3} & \cellcolor[HTML]{EFEFEF}\textbf{12.9} & 0.0 & 0.0 \\
 & Learn T5 & 41.4 & 61.7 & 0.2 & 0.1 & 91.4 & 87.8 & 0.3 & 0.1 & \cellcolor[HTML]{EFEFEF}\textbf{97.9} & \cellcolor[HTML]{EFEFEF}\textbf{98.8} \\
\midrule
Cross & Learn the first 4 tasks & \cellcolor[HTML]{EFEFEF}\textbf{61.8} & \cellcolor[HTML]{EFEFEF}\textbf{28.9} & \cellcolor[HTML]{EFEFEF}\textbf{3.0} & \cellcolor[HTML]{EFEFEF}\textbf{4.0} & \cellcolor[HTML]{EFEFEF}\textbf{3.5} & \cellcolor[HTML]{EFEFEF}\textbf{2.0} & \cellcolor[HTML]{EFEFEF}\textbf{95.9} & \cellcolor[HTML]{EFEFEF}\textbf{95.1} & 0.0 & 0.0 \\
Entropy  & Unlearn T2 & 98.2 & 99.2 & \cellcolor[HTML]{EFEFEF}\textbf{0.1} & \cellcolor[HTML]{EFEFEF}\textbf{0.01} & 0.1 & 0.1 & 7.6 & 11 & 0.0 & 0.0 \\
($\mathcal{L}_{ce}$) & Unlearn T4 & 94.2 & 0.2 & 0.1 & 0.1 & 0.1 & 0.3 & \cellcolor[HTML]{EFEFEF}\textbf{0.1} & \cellcolor[HTML]{EFEFEF}\textbf{0.1} & 85.4 & 90.3 \\
 & Learn T5 & 38 & 68.5 & 0.2 & 0.1 & 0.4 & 0.3 & 0.02 & 0.1 & \cellcolor[HTML]{EFEFEF}\textbf{0.1} & \cellcolor[HTML]{EFEFEF}\textbf{0.3} \\
\midrule
Supervised  & Learn the first 4 tasks & \cellcolor[HTML]{EFEFEF}\textbf{71.6} & \cellcolor[HTML]{EFEFEF}\textbf{75.0} & \cellcolor[HTML]{EFEFEF}\textbf{53.7} & \cellcolor[HTML]{EFEFEF}\textbf{38.8} & \cellcolor[HTML]{EFEFEF}\textbf{55.0} & \cellcolor[HTML]{EFEFEF}\textbf{64.2} & \cellcolor[HTML]{EFEFEF}\textbf{98.2} & \cellcolor[HTML]{EFEFEF}\textbf{97.6} & 0.0 & 0.0 \\
Contrastive & Unlearn T2 & 89.2 & 95.6 & \cellcolor[HTML]{EFEFEF}\textbf{0.3} & \cellcolor[HTML]{EFEFEF}\textbf{0.9} & 78.3 & 78.5 & 96.7 & 90.9 & 0.0 & 0.0 \\
Distillation  & Unlearn T4 & 88.6 & 94.7 & 0.1 & 0.2 & 97.8 & 94.7 & \cellcolor[HTML]{EFEFEF}\textbf{62.6} & \cellcolor[HTML]{EFEFEF}\textbf{19.1} & 0.0 & 0.0 \\
($\mathcal{L}_{scd}$) & Learn T5 & 48.4 & 32.2 & 0.2 & 0.2 & 90.6 & 88.2 & 0.1 & 0.3 & \cellcolor[HTML]{EFEFEF}\textbf{98.3} & \cellcolor[HTML]{EFEFEF}\textbf{99.0} \\
\bottomrule
\end{tabular}}
\caption{Ablative analysis of removing loss components in CL module and its impact across different execution sequences. CIFAR-10 with buffer size 500 in a $2\times5$ task distribution setup.}
\label{tab:loss_removal_comparison}
\end{table}


\subsection{Single task UL with $5\times2$ distribution} 
Single task unlearning in a $5\times2$ task distribution setting is given in Table~\ref{tab:cifar10_5x2}. The experiment explores the impact of different buffer sizes on performance of CL/UL execution sequence, which involves learning two Tasks and then unlearning Task 2. The results show the accuracy percentages for 10 tasks (C0 to C9) across different buffer sizes. As the buffer size increases, we observe a general improvement in performance, particularly in retaining accuracy for previously learned tasks. With a buffer size of 200, there's a noticeable drop in accuracy for task 2 (C1) after unlearning, while other tasks maintain relatively high accuracy. As the buffer size increases to 500 and then to 5120, the accuracy for task 2 after unlearning improves significantly (from $\sim0.0$ percent to 5.4 percent to 26.4 percent), indicating better retention of partial information. Additionally, the accuracy for other tasks generally improves or remains stable with larger buffer sizes, suggesting that increased buffer capacity allows for better preservation of knowledge across tasks during the continual learning and unlearning process.\par

\subsection{Comparison with CL and UL baselines} 
A comparison between the proposed method and a baseline UL (retrained) model, covering five tasks on the CIFAR-10 dataset, is presented in Table~\ref{tab:ul_comparison}. The results show that the proposed method generally performs similarly to or slightly below the baseline for most tasks. Notably, we see   significantly with Task 2 and Task 4 (columns C2, C3, C6, C7), showing $\sim0.0\%$ performance across all batch sizes. However, the proposed method demonstrates some improvements over the baseline in Task 3 (C5) and Task 5 (C8, C9), particularly at higher batch sizes. As the batch size increases, both methods tend to show improved performance across tasks, with the most substantial improvements seen in Task 1 and Task 3. A comparison of the proposed method with several baseline Continual Learning (CL) approaches, including DER, oEWC, and SGD, on the CIFAR-10 dataset is presented in Table \ref{tab:cl_comparison}. The results clearly demonstrate that the proposed method significantly outperforms the baseline CL methods, especially for retained classes. The performance of all methods improves as the batch size increases from 200 to 5120. Notably, the proposed method shows a marked improvement after unsupervised learning (UL) compared to before UL, particularly for retained classes. For instance, at a batch size of 5120, the proposed method achieves $93.9\%$ accuracy on retained classes after UL, compared to $89.5\%$ before UL, and substantially outperforms the next best method (DER) which achieves $83.81\%$. This suggests that the proposed method is particularly effective in maintaining performance on previously learned tasks while adapting to new information.\\

\subsection{Ablation Studies}
\textbf{Impact of Bernoulli probability.} The effect of varying Bernoulli Probability values (0.5, 0.6, and 0.7) on the performance of CIFAR-10 with a buffer size of 5120 is examined on Table \ref{tab:bp_ablation}. The execution sequence involves learning the first 4 tasks, then unlearning task 2 and task 4, followed by learning task 5. The results show accuracy percentages for 9 tasks (C1 to C9) under different probability settings. As the Bernoulli Probability increases from 0.5 to 0.7, we observe a general trend of improved accuracy across most tasks, especially for the tasks that were not unlearned (C1, C3, C5). Interestingly, the accuracy for unlearned tasks (C2 and C4) remains consistently low (0.0) across all probability values, indicating effective unlearning. The performance on task 5 (C5) improves notably as the probability increases, suggesting that higher Bernoulli Probability values might lead to better learning of new tasks after the unlearning phase. However, the impact on tasks 6-9 (C6-C9) is mixed, with some tasks showing improvement and others remaining relatively stable or slightly decreasing in accuracy as the probability increases.\par

\textbf{Impact of buffer size.} In this work, we analyze a system that performs both continual learning and unlearning using a replay/memory buffer. As new classes are learned, their examples are added to the buffer, and when unlearning is triggered, examples of classes that need to be forgotten are removed. However, as the buffer size increases, the system becomes better at continual learning but worse at unlearning. This is because a larger buffer helps retain knowledge from earlier tasks but also makes it harder to completely remove information related to classes that should be forgotten. In the Appendix, we provide a detailed mathematical analysis of this trade-off between learning and unlearning performance.

\textbf{Impact of different loss components in CL loss.} Table \ref{tab:loss_removal_comparison} provides an ablative analysis of removing loss components in the CL module and its impact on $2\times5$ task distribution setup. Across all execution sequences, the effect of loss removal varies significantly depending on the learning strategy. For initial task learning, online and contrastive loss show high accuracy across most tasks, while cross-entropy yields noticeably lower results, particularly for tasks C2-C5. Upon unlearning tasks (e.g., T2 and T4), accuracy for unlearned tasks drops to near-zero levels in most cases, confirming effective unlearning. This demonstrates that loss components influence both task-specific performance and knowledge retention in CL settings.



\section{Conclusion}
In this paper, we introduced a unified framework that addresses the dual challenges of continual learning and machine unlearning, which have traditionally been treated in isolation. By leveraging a controlled knowledge distillation mechanism, our approach enables models to selectively erase specific knowledge while continuing to learn new information, thus achieving a balance between learning and forgetting. We present a loss function tailored to handle both continual learning and unlearning tasks, even when requested in random sequences. Our experiments demonstrate that the framework effectively maintains performance in both tasks, providing a robust solution for dynamic environments where adaptability is essential. This work opens new avenues for developing more flexible and resilient machine learning models capable of meeting the demands of real-world applications.
{
\small
\bibliographystyle{ieeenat_fullname}
\bibliography{main.bib}
}

\newpage
\appendix

\newpage
\setcounter{page}{1}
\section{Appendix}

\subsection{Trade-off Between Continual Learning and Unlearning Performance}
In this section we explain why model performance improves for continual learning but degrades during unlearning phase on larger buffer size.
Let us consider a system that jointly performs continual learning and unlearning based on a replay buffer. In this setup:

\begin{itemize}
    \item Let \( \mathcal{B} \) be a replay buffer of size \( N \) that stores samples \( \{ x_1, x_2, \dots, x_N \} \), where each sample \( x_i \) is associated with a class label \( y_i \in \{ 1, 2, \dots, C \} \) for \( C \) classes.
    \item Let \( P_{\text{CL}}(N) \) represent the continual learning performance, and let \( E_{\text{UL}}(N) \) represent the unlearning effectiveness.
\end{itemize}

\textbf{Continual Learning Performance.} Continual learning performance improves as the replay buffer size \( N \) increases because more diverse samples from previous tasks are retained. We model the performance as a function of the buffer size:

\[
P_{\text{CL}}(N) = \alpha \log(N),
\]

where \( \alpha > 0 \) is a proportionality constant. The logarithmic relationship indicates that while increasing \( N \) improves performance, the gains diminish as \( N \) becomes large.

\textbf{Unlearning Effectiveness.} Unlearning requires the removal of samples belonging to a specific class \( y_u \) from the buffer. However, as \( N \) increases, it becomes harder to ensure that all related samples are removed. Let \( \rho(N) \) represent the probability that a sample of class \( y_u \) remains in the buffer after unlearning:

\[
\rho(N) = \frac{\beta}{N},
\]

where \( \beta > 0 \) is a constant. Thus, the unlearning effectiveness can be modeled as:

\[
E_{\text{UL}}(N) = 1 - \rho(N) = 1 - \frac{\beta}{N}.
\]

As \( N \) increases, \( E_{\text{UL}}(N) \) decreases, indicating worse unlearning performance.

\subsubsection{Mathematical Analysis of the Trade-Off Between Continual Learning and Unlearning}

The performance of a system engaging in both continual learning and unlearning tasks is determined by a trade-off between the effectiveness of continual learning, \(P_{\text{CL}}(N)\), and unlearning, \(E_{\text{UL}}(N)\). This relationship can be expressed as:

\[
\text{Total Performance}(N) = P_{\text{CL}}(N) \cdot E_{\text{UL}}(N),
\]

where \(N\) represents the buffer size, a critical parameter influencing both objectives. Substituting the functional forms of \(P_{\text{CL}}(N)\) and \(E_{\text{UL}}(N)\), we define:

\[
P_{\text{CL}}(N) = \alpha \log(N), \quad E_{\text{UL}}(N) = 1 - \frac{\beta}{N},
\]

where:
\begin{itemize}
    \item \(\alpha > 0\): Scaling factor for the contribution of the buffer to continual learning,
    \item \(\beta > 0\): Sensitivity factor for unlearning effectiveness with respect to buffer size.
\end{itemize}

This leads to the explicit expression for total performance:

\[
\text{Total Performance}(N) = \alpha \log(N) \left(1 - \frac{\beta}{N}\right).
\]

\textbf{Behavior of Continual Learning and Unlearning Components}

\paragraph{1. Continual Learning Term (\(P_{\text{CL}}(N)\))}  
The term \(P_{\text{CL}}(N) = \alpha \log(N)\) grows logarithmically as \(N\) increases. This reflects the improvement in continual learning performance due to the increased ability of the system to retain diverse and sufficient information for generalization.

\[
\lim_{N \to \infty} P_{\text{CL}}(N) = \infty.
\]

\paragraph{2. Unlearning Effectiveness Term (\(E_{\text{UL}}(N)\))}  
The term \(E_{\text{UL}}(N) = 1 - \frac{\beta}{N}\) asymptotically approaches 1 as \(N\) increases. For finite \(N\), unlearning effectiveness decreases as \(\frac{\beta}{N}\) dominates for smaller buffer sizes. However, the marginal improvement in \(E_{\text{UL}}(N)\) diminishes as \(N\) grows:

\[
\lim_{N \to \infty} E_{\text{UL}}(N) = 1.
\]

\textbf{Total Performance Behavior}

The total performance function is given by:

\[
\text{Total Performance}(N) = \alpha \log(N) \left(1 - \frac{\beta}{N}\right).
\]

Analyzing its derivative with respect to \(N\) provides insight into the interplay between continual learning and unlearning:

\[
\frac{d}{dN} \text{Total Performance}(N) = \alpha \left(\frac{1}{N} - \frac{\beta}{N^2} - \frac{\beta \log(N)}{N^2}\right).
\]

The first term, \(\frac{1}{N}\), reflects the logarithmic growth of continual learning performance. The second and third terms, \(-\frac{\beta}{N^2}\) and \(-\frac{\beta \log(N)}{N^2}\), capture the diminishing contribution of unlearning effectiveness and its interplay with buffer size. Setting the derivative to zero:

\[
\frac{d}{dN} \text{Total Performance}(N) = 0,
\]

yields the critical point \(N^*\), representing the buffer size at which the trade-off between continual learning and unlearning is balanced. Solving for \(N^*\) involves finding the root of the equation:

\[
\frac{1}{N^*} = \frac{\beta}{(N^*)^2} + \frac{\beta \log(N^*)}{(N^*)^2}.
\]

This transcendental equation can be solved numerically for specific values of \(\alpha\) and \(\beta\), providing the buffer size that optimally balances the trade-off.

\textbf{Asymptotic Behavior}

In the limit as \(N \to \infty\), the behavior of the total performance is dominated by the growth of \(P_{\text{CL}}(N)\), while \(E_{\text{UL}}(N)\) approaches its asymptotic value:

\[
\lim_{N \to \infty} \text{Total Performance}(N) = \alpha \log(N).
\]

This result indicates that for extremely large buffers, continual learning performance becomes the primary contributor to total performance, and the unlearning term \(E_{\text{UL}}(N)\) stabilizes without significantly impacting the total performance.

\textbf{Interplay of \(\alpha\) and \(\beta\)}

The parameters \(\alpha\) and \(\beta\) govern the relative importance of continual learning and unlearning:
\begin{itemize}
    \item Larger \(\alpha\) values amplify the contribution of continual learning to the total performance.
    \item Larger \(\beta\) values increase the sensitivity of unlearning effectiveness to buffer size, emphasizing its role in smaller buffers.
\end{itemize}

These parameters provide a means of tuning the system's priorities, and the resulting performance curve can exhibit varying shapes depending on the relative values of \(\alpha\) and \(\beta\).

\begin{algorithm}[t]
\caption{CL-UL (Continual Learning and Unlearning)}
\label{alg:cl_ul}
\textbf{Parameters:} 
\begin{itemize}
    \item \textbf{Teacher parameters:} $\Theta_T$, $\Psi_T$, $\Phi_T$
    \item \textbf{Bad teacher parameters:} $\Theta_b$, $\Psi_b$, $\Phi_b$
    \item \textbf{Student parameters:} $\Theta_s$, $\Psi_s$, $\Phi_s$
    \item \textbf{Item label:} $y$
    \item \textbf{Hyperparameters:} $\alpha_{ul}$ (learning-unlearning weighting), $\alpha_1, \alpha_2, \alpha_3$ (loss coefficients), $\eta$ (learning rate)
\end{itemize}
\textbf{Initialization:} 
\begin{itemize}
    \item Buffer $\mathcal{B} \gets \{\}$ (empty buffer)
    \item Stream Data $D = \bigcup\limits_{i=1}^{T} D_i$
\end{itemize}

\begin{algorithmic}[1]
\FOR{$t \in \{1, 2, \ldots, T\}$}
    \STATE Initialize the loss $\mathcal{L} \gets 0$
    
    \STATE Compute the primary task loss:
    $\begin{aligned}
\mathcal{L}_{task} = & \ \alpha_{ul} \cdot \mathrm{cross\_entropy}\left(f_{\Theta_s, \Phi_s}(x), y\right) \\
& + (1 - \alpha_{ul}) \cdot \mathrm{KL\_div}\left(f_{\Theta_b, \Phi_b}(x), f_{\Theta_s, \Phi_s}(x)\right)
\end{aligned}$
    
    \STATE Sample from the buffer: $(X_B, Y_B) \gets \mathrm{Sample}(\mathcal{B})$
    
    \STATE Calculate auxiliary losses:
    \begin{itemize}
        \item $\mathcal{L}_{od}$ (out-of-distribution loss) using Eq. (4)
        \item $\mathcal{L}_{cd}$ (class-discrimination loss) using Eq. (5)
        \item $\mathcal{L}_{scd}$ (sample-consistency discrimination loss) using Eq. (7)
    \end{itemize}
    
    \STATE Aggregate the losses:
    \[
    \mathcal{L} \gets \mathcal{L}_{task} + \alpha_1 \cdot \mathcal{L}_{od} + \alpha_2 \cdot \mathcal{L}_{cd} + \alpha_3 \cdot \mathcal{L}_{scd}
    \]
    
    \STATE Update student parameters:
    \[
    (\Theta_s, \Psi_s, \Phi_s) \gets (\Theta_s, \Psi_s, \Phi_s) - \eta \cdot \frac{\partial \mathcal{L}}{\partial (\Theta_s, \Psi_s, \Phi_s)}
    \]
    
    \STATE Update teacher parameters with random momentum:
    \[
    (\Theta_T, \Psi_T, \Phi_T) \gets \mathrm{MomentumUpdate}((\Theta_T, \Psi_T, \Phi_T))
    \]
    
    \STATE Optionally update buffer $\mathcal{B}$ with new samples.
\ENDFOR
\end{algorithmic}
\end{algorithm}

\subsection{Related Work}
\textbf{Continual Learning.} Continual learning approaches can be broadly categorized into five types: regularization-based, replay-based, optimization-based, representation-based, and architecture-based \cite{wang2024comprehensive}. Regularization-based methods add terms to balance the learning of new and old tasks, either by regularizing weight deviations \cite{kirkpatrick2017overcoming,ritter2018online,roady2020stream} or by ensuring output closeness on old data, which is often substituted with new samples \cite{dhar2019learning,iscen2020memory} or synthetic data \cite{wu2018memory,zhai2019lifelong}. Replay-based methods use memory buffers to store and replay subsets of old data, with strategies to optimize buffer use \cite{chaudhry2019tiny,prabhu2020gdumb}. Some methods employ feature-based replay for efficiency and privacy \cite{liu2020generative}. Optimization-based approaches adjust optimization processes, such as gradient projections in GEM \cite{lopez2017gradient} or pathfinding in MC-SGD \cite{mirzadeh2020linear}. Representation-based methods leverage stable model representations, like LUMP \cite{madaan2021representational}, which interpolates between tasks, or DualNet \cite{pham2021dualnet}, which uses a dual learner strategy. Architecture-based methods involve specific architectural designs, allocating parameters from a fixed or dynamic budget \cite{mallya2018piggyback, yoon2017lifelong}, or mimicking biological mechanisms \cite{ororbia2020continual, ororbia2022lifelong}. Some methods separate parameters for task-shared and task-specific learning \cite{ebrahimi2020adversarial}.

\textbf{Machine Unlearning.} Cao et al. \cite{cao2015towards} first introduced machine unlearning by converting learning algorithms into a summation form, allowing selective deletion of terms when unlearning requests were made. This method worked for simple algorithms like SVMs and Naive Bayes. Later, Bourtoule et al. \cite{bourtoule2021machine} developed SISA, an ensemble-based framework that facilitated unlearning by training smaller models on non-overlapping data subsets, with unlearning applied only to the relevant models. Machine Unlearning can be classified into \textit{Exact Unlearning}, like SISA, which ensures models are indistinguishable from retrained versions, and \textit{Approximate Unlearning}, as proposed by Ginart et al. \cite{ginart2019making}, which produces models similar, but not identical, to retrained ones.

In contrast to traditional algorithms like Bayes, SVMs, K-means, and Random Forests, deep learning unlearning primarily follows the probabilistic approach due to its non-convex nature. For instance, \cite{golatkar2020eternal} improved unlearning efficiency using the Fisher Information Matrix (FIM), and \cite{golatkar2020forgetting} enhanced this with NTK theory. Mehta et al. \cite{mehta2022deep} ensured unlearning in large networks by targeting key parameters before Hessian calculation, while Tarun et al. \cite{tarun2023fast} used noise samples to fine-tune and unlearn. Teacher-student frameworks have also been explored, as in \cite{chundawat2023can}, which uses separate teachers for retaining and forgetting information.~\cite{foster2024fast} employed FIM to determine the importance of each parameter and subsequently reduced the influence of parameters crucial to the forget set during inference. \cite{foster2024loss} extended \cite{foster2024fast} by replacing the diagonal of the FIM with the gradient of the $l_2$ norm of the model outputs, resulting in an unlearning algorithm that neither requires labels nor loss calculations.

Recent explorations in generative models include unlearning concepts and objects from diffusion models, such as minimizing attention map norms \cite{zhang2023forget}, and applying unlearning layers in Large Language Models (LLMs) \cite{chen2023unlearn}. Model-editing approaches, like those in \cite{patil2023can}, have also been used to unlearn sensitive information from LLMs.

\textbf{Knowledge Distillation.} Knowledge Distillation (KD) is integral to replay-based Continual Learning (CL) methods. Typically, the model trained on previous tasks acts as the teacher, while the model learning new tasks serves as the student, with enforced similarity in intermediate or final layer outputs. A challenge in CL is the lack of older training samples. To address this, methods like \cite{iscen2020memory} use new training images for KD, learning a Feature Adaptation Network to bridge features from both the old and new models. Other approaches, such as \cite{douillard2020podnet}, employ a small subset of older samples to calculate an efficient spatial distillation loss, enabling CL across many episodes. Additionally, works like \cite{wu2018memory} and \cite{zhai2019lifelong} use synthetic data for KD.

KD is also utilized in Machine Unlearning (MU). For instance, \cite{chundawat2023zero} generates informative synthetic samples, with the original model as the teacher and the unlearning model as the student, filtering the forget set with a band-pass filter. \cite{chundawat2023can} employs a two-teacher approach, using the original model to distill knowledge from the retain set and a randomly initialized teacher for misinformation on the forget set. \cite{kurmanji2024towards} introduces a single-teacher method where the student learns to selectively disobey the teacher.

Our work unifies Continual Learning and Machine Unlearning through knowledge distillation. We employ contrastive KD to mitigate catastrophic forgetting, adaptive distillation for new task learning, and an unlearning teacher to distill misinformation about tasks to forget. The working of the proposed method is presented in Algorithm~\ref{alg:cl_ul}.
\appendix
\end{document}